\documentclass[final,3p,times,twocolumn,authoryear]{elsarticle}
\usepackage{amssymb}
\usepackage{amsmath}
\usepackage{graphicx}
\usepackage{array}
\usepackage{makecell} 
\usepackage{booktabs} 
\usepackage{amsmath} 
\usepackage{amssymb} 
\usepackage{hyperref} 
\usepackage{booktabs}
\usepackage{adjustbox}
\usepackage{multirow}
\usepackage{caption}

\begin{document}
\begin{frontmatter}
\title{A Language-Signal-Vision Multimodal Framework for Multitask\\ Cardiac Analysis} 

\author[a]{Yuting Zhang}
\author[a]{Tiantian Geng}
\author[a]{Luoying Hao}
\author[a]{Xinxing Cheng}
\author[a]{Alexander Thorley}
\author[b,c]{Xiaoxia Wang}
\author[d]{Wenqi Lu}
\author[e]{Sandeep S Hothi}
\author[f]{Lei Wei\corref{cor1}}
\author[g]{Zhaowen Qiu}
\author[b,c,h]{Dipak Kotecha\corref{cor1}} 
\author[a,i]{Jinming Duan\corref{cor1}}

\address[a]{School of Computer Science, University of Birmingham, Birmingham, UK}
\address[b]{Department of Cardiovascular Sciences, University of Birmingham, Birmingham, UK}
\address[c]{NIHR Birmingham Biomedical Research Centre and West Midlands NHS Secure Data Environment, University Hospitals Birmingham NHS Foundation Trust, Birmingham, UK}
\address[d]{Department of Computing and Mathematics, Manchester Metropolitan University, Manchester, UK}
\address[e]{Department of Cardiology, Heart and Lung Centre, Royal Wolverhampton NHS Trust, Wolverhampton, UK}
\address[f]{Department of Cardiovascular Surgery, The First Affiliated Hospital with Nanjing Medical University , Nanjing,China}
\address[g]{College of Computer and Control Engineering, Northeast Forestry University, Harbin, China}
\address[h]{Julius Center, University Medical Center Utrecht, the Netherlands}
\address[i]{Division of Informatics, Imaging and Data Sciences, University of Manchester, Manchester, UK}

\begin{abstract}
Contemporary cardiovascular management involves complex consideration and integration of multimodal cardiac datasets, where each modality provides distinct but complementary physiological characteristics. While the effective integration of multiple modalities could yield a holistic clinical profile that accurately models the true clinical situation with respect to data modalities and their relatives weightings, current methodologies remain limited by: 1) the scarcity of patient- and time-aligned multimodal data; 2) reliance on isolated single-modality or rigid multimodal input combinations; 3) alignment strategies that prioritize cross-modal similarity over complementarity; and 4) a narrow single-task focus. In response to these limitations, a comprehensive multimodal dataset was curated for immediate application, integrating laboratory test results, electrocardiograms, and echocardiograms with clinical outcomes. Subsequently, a unified framework, Textual Guidance Multimodal fusion for Multiple cardiac tasks (TGMM), was proposed. TGMM incorporated three key components: 1) a MedFlexFusion module designed to capture the unique and complementary characteristics of medical modalities and dynamically integrate data from diverse cardiac sources and their combinations; 2) a textual guidance module to derive task-relevant representations tailored to diverse clinical objectives, including heart disease diagnosis, risk stratification and information retrieval; and 3) a response module to produce final decisions for all these tasks. Furthermore, this study systematically explored key features across multiple modalities and elucidated their synergistic contributions in clinical decision-making. Extensive experiments showed that TGMM outperformed state-of-the-art methods across multiple clinical tasks, with additional validation confirming its robustness on another public dataset. The code and trained models will be publicly available at \url{https://github.com/ytz300/TGMM}.
\end{abstract}

\begin{keyword}
  Multimodal learning \sep Multi-task learning \sep Foundation models \sep Heart failure diagnosis \sep Risk stratification \sep Cardiovascular disease  \sep Model explanation 
\end{keyword}
\end{frontmatter}

\section{Introduction}
Technological advancements over recent decades have dramatically transformed the healthcare landscape, leading to a growing diversity of quantitative and qualitative clinical data types for diagnosis and patient care ~\cite{weintraub2019role}. This transformation has been particularly profound in the management of complex chronic conditions~\cite{heidenreich20222022, topol2019high}. Heart failure (HF) stands out as a prototypical example, given its clinical complexity and enhanced insights provided by multi-dimensional data streams, such as clinical history, laboratory tests, electrocardiograms (ECGs), and cardiac ultrasound (echocardiogram) ~\cite{heidenreich20222022b,heidenreich20222022}. However, most artificial intelligence (AI) applications in cardiovascular medicine are still restricted by a reliance on single-modal or rigid multimodal inputs, typically focusing on a specific task~\cite{banerjee2023identifying, akerman2025external, zhang2024development}. This stands in contrast to the integrative approach used in clinical practice, where clinicians dynamically combine information from multiple modalities for diagnosis, prognostication, and treatment decisions. The gap between clinical and AI decision-making reinforces the limitations of current machine learning methods in replicating real-world clinical reasoning and/or providing clinical utility ~\cite{gill2023artificial}.

While multimodal learning (MML) techniques have achieved notable progress at the algorithmic and methodological level, applying them effectively in clinical medicine remains a nontrivial challenge. For example, although traditional methods have shown the ability to integrate features from individual modalities, they often lack filtering mechanisms to distinguish clinically relevant information and introduce noise, particularly when dealing with highly heterogeneous medical data~\cite{wolf2022daft, acosta2022multimodal}. Attention-based methods ~\cite{chen2023medical, mo2024unveiling, shan2024contrastive, alayrac2022flamingo} and loss-constrained methods ~\cite{he2020momentum, bansal2023cleanclip, li2021align, christensen2024vision} methods have been proposed as alternatives, however, they primarily focus on querying similar information from another modality or aligning features in high-dimensional space. This neglects to account for the shared and complementary information encompassed within medical data. For instance, echocardiograms and laboratory tests can both assess cardiac function, yet they also provide unique insights: imaging identifies structural and functional abnormalities, while laboratory tests such as B-type natriuretic peptide reflect overall cardiac strain.

Current MML techniques that rely solely on shared information might be insufficient to capture the full spectrum of clinical multimodal data. In addition, medical data inherently support multiple diagnostic tasks, with the same dataset applicable to various purposes. For example, echocardiograms can aid in both HF diagnosis and in prognostic stratification, although the critical regions of interest differ based on the specific clinical task ~\cite{akerman2025external}. With the advancement of large multimodal models like Qwen2.5-VL~\cite{wang2024qwen2}, PaLM-E~\cite{driess2023palm} and Janus-Pro~\cite{chen2025janus}, prompt engineering has shown the potential of these unified architectures with billions of parameters to focus on multi-task scenarios~\cite{ye2023prompt}.  However, these large models lack domain-specific medical knowledge, and their fine-tuning demands substantial high-quality data and computational resources.

This challenge is particularly evident in the cardiovascular domain, where high-quality multimodal public data remain scarce. While national studies like the UK Biobank have collected diverse data such as ECGs and cardiovascular magnetic resonance imaging (CMR), the echocardiogram is notably absent despite its central role in cardiology~\cite{littlejohns2020uk, woodfield2015accuracy}. In contrast, institutional datasets like the Medical Information Mart for Intensive Care (MIMIC) offer rich cardiovascular data, such as MIMIC-IV ~\cite{johnson2024mimic}; however, these resources are often fragmented and lack integration at both the patient and temporal levels. Additionally, outcomes are labelled with administrative codes, such as the International Classification of Diseases (ICD) ~\cite{hong2023international}, which require specialized medical expertise for accurate interpretation. This complexity further restricts the usability of MIMIC datasets for multimodal machine learning applications.

Finally, explainability of model predictions is crucial in medical applications, as reliance on black-box models undermines clinical trust and hinders healthcare adoption~\cite{lundberg2017unified, tjoa2020survey}. Although numerous explanation methods have been developed, such as SHapley Additive exPlanations (SHAP)~\cite{lundberg2017unified} and Class Activation Mapping (CAM)~\cite{selvaraju2017grad}, most are applied in unimodal contexts or explain only individual modalities within multimodal models~\cite{christensen2024vision, ghorbani2020deep}. As a result, there is insufficient understanding of how different modalities interact to derive predictions and how these interactions yield valuable insights for clinicians. For instance, the explanation analysis conducted in this study revealed that, in high-risk HF prediction, the model placed particular emphasis on the T-wave in the ECG, in conjunction with other modalities (see Figure~\ref{fig:explain}(c)). However, without a comprehensive explanation of this decision process, it remains unclear whether this emphasis played a decisive role or aligned with established clinical knowledge.

In this study, a comprehensive multimodal dataset, HFTri-MIMIC, was curated from MIMIC-IV ~\cite{johnson2024mimic}. HFTri-MIMIC encompassed patient- and time-aligned laboratory tests (Labs), 12-lead ECGs, and echocardiograms (ECHOs) from 1,524 patients with prevalent HF (patients meeting criteria for HF) and non-prevalent HF (patients not meeting HF criteria), along with detailed follow-up information for risk stratification in 455 individuals. Unlike most public datasets~\cite{ouyang2020video,reddy2023video,leclerc2019deep}, which consist of preprocessed data (e.g., selecting representative apical four-chamber (A4C) views from raw DICOM files), this dataset remained in their raw formats, allowing the end-to-end model to learn directly from the original multimodal inputs without the need for manual curation or selection. Consequently, this dataset provides a strong foundation for the development of AI models capable of leveraging multimodal data in real-world applications.

With this dataset, a unified Textual Guidance Multimodal fusion framework was proposed to handle Multiple clinical cardiac tasks (TGMM), capable of dynamically and seamlessly integrating information from diverse cardiac modalities. Specifically, a modality-aware representation learning (MARL) module was employed to extract modality-specific features. To capture the unique and complementary characteristics of medical modalities, a MedFlexFusion module (MFFM) was then introduced to learn shared representations, while dynamically integrating data from diverse sources and their combinations. To further capture modality-specific and shared information, these were subsequently coupled and refined with semantic features derived from a textual guidance module (TGM) to extract task-relevant representations, which were then processed by a response module for task-specific prediction. By defining different semantic information in the TGM, this framework supported multiple downstream cardiac tasks, including disease diagnosis, prognosis, and information retrieval.

Experiments were then conducted to evaluate the TGMM framework using both a curated HFTri-MIMIC dataset and a public myocardial infarction (MI) dataset ~\cite{wagner2020ptb}. The results showed that trimodal integration of Labs, ECGs, and ECHOs improved HF diagnosis and risk prediction by up to 10\% and 8\%, respectively, while bimodal with Labs and ECGs improved MI diagnosis by up to 5\% over single modalities. Beyond its predictive performance, TGMM showed strong resilience to incomplete clinical data, consistently achieving reliable results across diverse modality combination scenarios. Information retrieval tasks were also experimented with to show its potential for multi-task applications. To further assess model reliability, ablation studies confirmed the contribution of each component. Furthermore, explainability analysis highlighted the importance of both modality-specific and shared features in clinical tasks. The decision-making basis of task-specific outputs was also systematically explored by analyzing key features across multiple modalities and elucidating their synergistic contributions. In summary, the key contributions can be articulated as follows:
\begin{itemize}    
    \item A comprehensive multimodal dataset, HFTri-MIMIC, was curated from public sources with patient- and time-aligned lab tests, 12-lead ECGs, raw echocardiograms, and clinical outcomes, providing a ready-to-use foundation for developing models in real-world multimodal clinical settings.
    \item An end-to-end framework was proposed, designed to handle trimodal, bimodal, and unimodal data for multi-task prediction (e.g., diagnosis, risk stratification), aligning with clinical practices that combine diverse modalities to support diagnosis and treatment decisions.
    \item Extensive experiments showed high predictive accuracy in all trimodal, bimodal, and unimodal configurations across multiple downstream tasks, with extra validation on another public dataset further confirming the effectiveness of the proposed method.
    \item Explainable AI techniques were employed in a multimodal configuration to provide transparent feature attributions and to illustrate the joint contributions of each modality to predictive reasoning, thereby fostering clinical trust and supporting the future clinical validation of multimodal AI systems.
\end{itemize}

\section{Related Work}
\subsection{Cardiovascular Datasets}
Current publicly available cardiovascular datasets predominantly feature single-modality data. For example, PTB-XL ~\cite{wagner2020ptb} and MIT-BIH~\cite{moody2001impact} contain only ECG signals, while EchoNet-Dynamic ~\cite{ouyang2020video}, EchoNet-Pediatric ~\cite{reddy2023video}, and CAMUS ~\cite{leclerc2019deep} offer only processed echocardiograms (e.g., A4C or a few standardized views). While large-scale population studies such as the UK Biobank include diverse clinical data (e.g., ECGs and CMR), echocardiograms are notably absent ~\cite{littlejohns2020uk, woodfield2015accuracy}. Institutional datasets like MIMIC-IV \cite{johnson2024mimic} offer multimodal data but suffer from fragmentation, lack cohesive alignment at patient and time levels, and rely on outcomes derived from administrative codes \cite{hong2023international} that require specialized clinical expertise for accurate interpretation. These challenges limit the immediate applicability of these resources to multimodal machine learning applications. In this study, a well-aligned multimodal dataset was curated to enable efficient AI model development.

\subsection{Representation Learning}
Supervised learning (SL) has traditionally enabled representation learning with labelled data, achieving notable success in natural language processing (NLP)~\cite{raffel2020exploring}, computer vision (CV)~\cite{liu2021deep}, and signal processing~\cite{sherly2023efficient}. Recently, self-supervised learning (SSL) has emerged as a powerful paradigm for general-purpose representation learning, forming the basis for foundation models. These models are usually pre-trained with generative learning (e.g., next-token prediction~\cite{gu2021domain}, masked-based modelling ~\cite{naguiding, geng2022multimodal}), contrastive learning~\cite{huang2023visual, pai2024foundation, wang2024pathology, vukadinovic2024echoprime}, or hybrid learning~\cite{huang2024enhancing, lu2024visual, vorontsov2024foundation, xu2024whole}. By leveraging diverse and large-scale unlabelled datasets, domain-specific foundation models such as Clinical-BERT~\cite{yan2022clinical}, BiomedGPT~\cite{zhang2024generalist}, MedCPT~\cite{jin2023medcpt}, ST-MEM~\cite{naguiding} have been developed for specialized medical applications. In light of the limited availability of large-scale cardiac multimodal datasets, this study employed foundation models to enable modality-aware learning.

\subsection{Multimodal Learning}
Multimodal learning integrates representations across different modalities, typically involving feature extraction followed by fusion. Existing fusion techniques include traditional merge-operation-based~\cite{vale2021long}, uncertainty-based~\cite{mo2024unveiling} and attention-based~\cite{zhou2023cross}. These techniques either lack mechanisms to filter clinically relevant information or primarily focus on feature alignment, overlooking the complementary nature inherent in medical data. SSL also is an effective method in exploring relationships between modalities, contributing to multimodal foundation models like EchoCLIP~\cite{christensen2024vision} and LoVT~\cite{muller2022joint}, though these models require adaptation for downstream tasks. Recent unified models like Qwen2.5-VL~\cite{wang2024qwen2}, PaLM-E~\cite{driess2023palm}, and Janus-Pro~\cite{chen2025janus} offer progress, but they lack domain-specific medical knowledge and require substantial data and computational resources for fine-tuning. Additionally, their lack of transparency hinders clinical deployment~\cite{world2024ethics}.

\subsection{Explainable Artificial Intelligence (XAI)}
The explainability of model predictions is critical in medical applications, where reliance on black-box models undermines clinical trust and hinders adoption in practice~\cite{lundberg2017unified, tjoa2020survey}. While popular methods such as SHAP~\cite{lundberg2017unified} and CAM~\cite{selvaraju2017grad} have been widely used to interpret unimodal models, their extension to multimodal settings remains underexplored. For instance, existing research typically explains only one modality in isolation~\cite{christensen2024vision, ghorbani2020deep}, overlooking the interaction between multiple modalities in generating joint predictions. This is an aspect crucial for understanding integrated clinical reasoning in the context of multimodal learning. In this study, explainable AI techniques were employed under a trimodal configuration to illustrate transparent feature attributions across modalities.
\begin{figure}[t!]
    \centering
\includegraphics[width=1\linewidth]{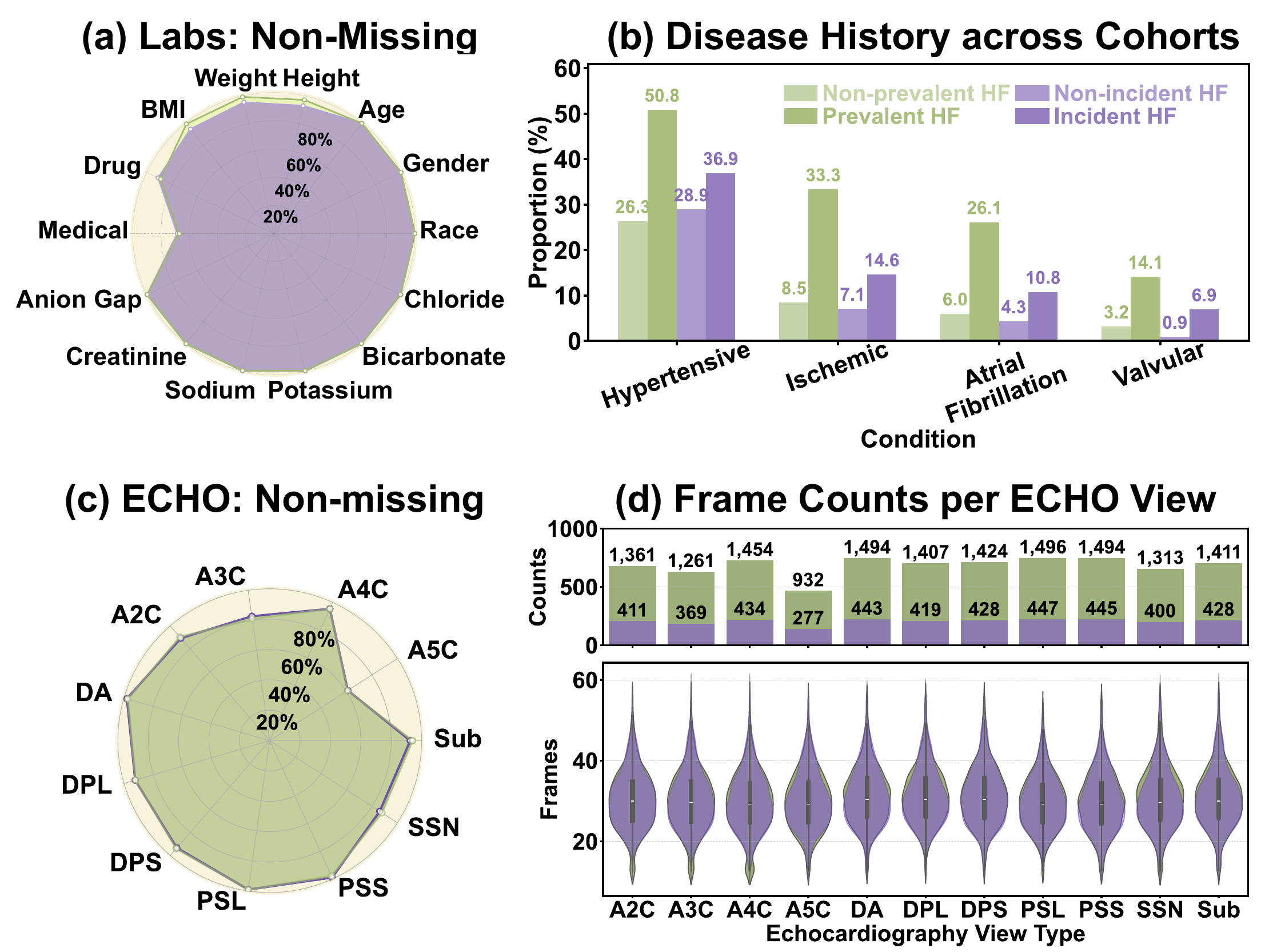}
    \caption{Illustrations of statistics for the diagnosis set (green) and the prognosis subset (purple) within the dataset: (a)Distribution of non-missing values for laboratory test results and metadata, where ``Medical'' and ``Drug'' refer to medical history and medication history, respectively. (b)Percentage of patients with different disease histories across various cohorts. (c) Distribution of non-missing views for echocardiograms. (d) Distribution of frames across echocardiogram views and their respective counts.}
    \label{fig:data}
\end{figure}
\section{HFTri-MIMIC Dataset}
\subsection{Overview}
The HFTri-MIMIC was sourced from the ethically approved MIMIC-IV, a deidentified dataset from Beth Israel Deaconess Medical Center (2008–2022)~\cite{johnson2023mimic}.  It consisted of three modalities: (1) laboratory test results (Labs), (2) 12-lead electrocardiograms (ECGs), and (3) echocardiograms (ECHOs). These modalities were patient- and time-aligned for 1,524 patients with either prevalent or non-prevalent heart failure (diagnosis set), with a subset of 455 individuals providing detailed follow-up information for risk stratification (prognosis subset). 
\subsection{Dataset Construction}
MIMIC-IV combines data from two hospital systems: a hospital-wide electronic health record (EHR) and an ICU-specific clinical information system, while HFTri-MIMIC focused exclusively on the hospital-wide EHR, primarily encompassing hospital admission records. For HFTri-MIMIC, records from the same hospital stay that included Labs, ECGs, and ECHOs were selected to ensure patient alignment. Additional exclusion criteria were applied to guarantee the collection of all three modalities within the same admission period, therefore ensuring temporal alignment and resulting in 1,538 samples. After excluding ten ECGs and four ECHOs with anomalies, the final dataset consisted of 1,524 samples, including 708 patients with heart failure (prevalent HF) and 816 patients without (non-prevalent HF), classified based on ICD-9/10 codes (see Table~\ref{tab:data}).  Furthermore, the follow-up cohort included clinical data linked to future health outcomes, particularly the development of HF. Among the 816 non-prevalent HF patients, follow-up information was available for 455 (29.9\%) individuals. Of these, 130 patients (28.6\%) developed incident HF, while 325 (71.4\%) experienced non-incident HF after more than a month (see Table~\ref{tab:data}). More details of the exclusion criteria and cohort construction were provided in Appendix A.
\subsection{Statistical Analysis}
Both the diagnosis and prognosis cohorts included Labs, ECGs, and ECHOs, with detailed patient characteristics provided in Appendix Table A.1. The Labs covered key biomarkers such as ``Anion Gap'', ``Bicarbonate'', ``Creatinine'', ``Potassium'', and ``Sodium'', along with additional metadata, including age, ethnicity, gender, medical history, drug history, body mass index (BMI), height, and weight. The majority of these features have over 90\% of values available, whereas medical history has the lowest availability (around 70\%, see Fig.~\ref{fig:data}(a)). Further analysis of historical data indicated that hypertensive diseases were prevalent across all patient cohorts (see Fig.~\ref{fig:data}(b)). For ECGs, they were uniformly 12-lead recordings, each with a duration of 10 seconds and sampled at a frequency of 500 Hz. Meanwhile, the ECHOs were acquired with GE Vivid machines and stored in DICOM format. To standardize the analysis, a view classifier \cite{christensen2024vision} was employed to automatically select 11 distinct views, comprising apical two-chamber (A2C), apical three-chamber (A3C), apical four-chamber (A4C), apical five-chamber (A5C), colour Doppler A4C (DA), colour Doppler parasternal long-axis (DPL), colour Doppler parasternal short-Axis (DPS), parasternal long-axis (PSL), parasternal short-axis (PSS), suprasternal short-axis (SSN), and subcostal view (Sub), from the raw DICOM. The results showed that only around 62\% of the samples included the A5C view, which had the highest missing rate (see Fig.\ref{fig:data}(c)), while Fig.\ref{fig:data}(d) illustrated the distribution of frames across these views, with most views containing approximately 30 frames each.
\begin{table}
  \centering
  \caption{Distribution of Heart Failure Cases Across Study Cohorts}
  \footnotesize  
  \label{tab:data}
  \setlength{\tabcolsep}{1.5pt}  
  \renewcommand{\arraystretch}{0.9}  
  \begin{tabular}{@{}cccc@{}} 
    \toprule
    \multicolumn{2}{c}{\textbf{Diagnosis (N=1,524)}} & \multicolumn{2}{c}{\textbf{Prognosis (N=455)}} \\
    \cmidrule(lr){1-2} \cmidrule(lr){3-4}
    \textbf{Prevalent HF} & \textbf{Non-Prevalent HF} & \textbf{Incident HF} & \textbf{Non-Incident HF} \\
    \midrule
    708 (46.5\%) & 816 (53.5\%) & 130 (29.9\%) & 325 (28.6\%) \\
    \bottomrule
    \multicolumn{4}{l}{\scriptsize{Note: HF = Heart Failure; N = Number of Patients.}} \\
  \end{tabular}
\end{table}
\section{Methodology}
\begin{figure*}
    \centering
    \includegraphics[width=1\linewidth]{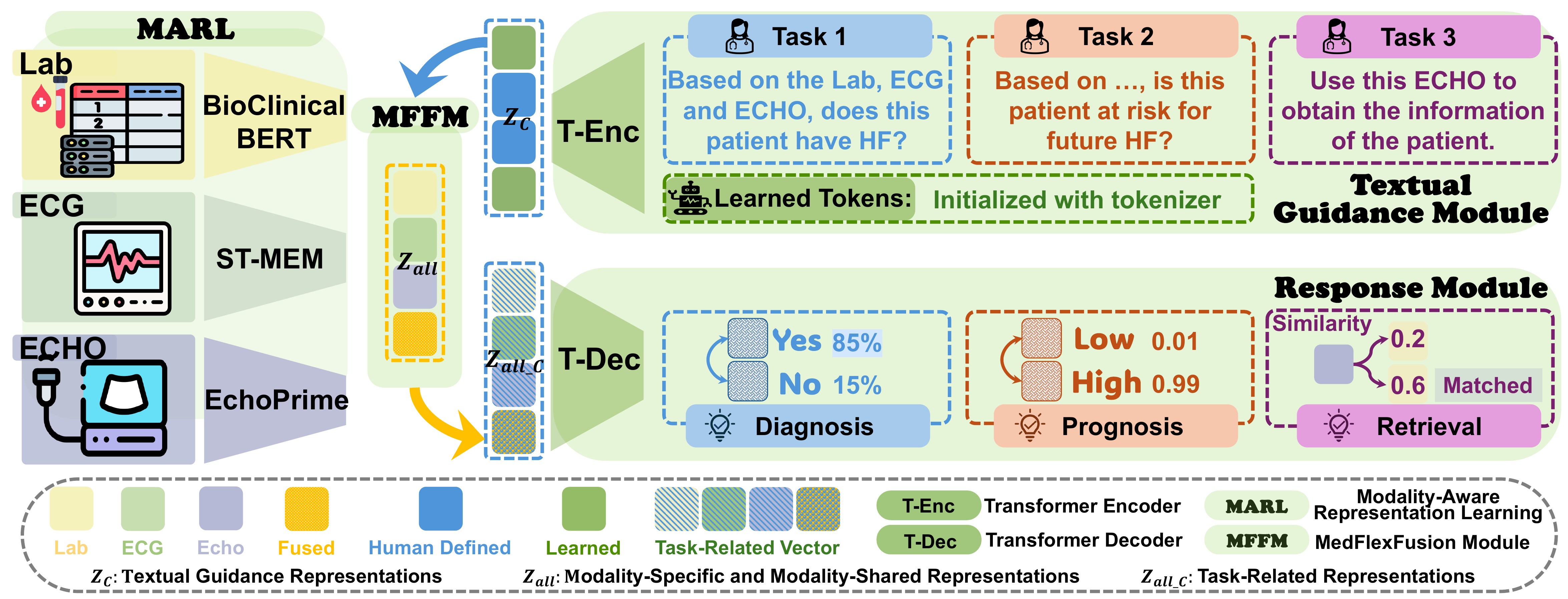}
    \caption{Illustration of Textual Guidance Multimodal fusion framework for Multiple Tasks (TGMM).}
    \label{fig:workflow}
\end{figure*}

The proposed TGMM framework comprised four principal components: modality-aware representation learning (MARL), MedFlexFusion module (MFFM), textual guidance module (TGM) and response module. MARL employed foundation models to extract modality-specific features, which were subsequently fused by the proposed MFFM for shared representations. Then, these shared and specific features were further coupled and refined with the TGM, enabling adaptive focus on task-relevant information, which facilitated task-specific predictions via the response module. An overview of the framework was provided in Fig.~\ref{fig:workflow}.

\begin{figure}[t!]
    \centering
    \includegraphics[width=1\linewidth]{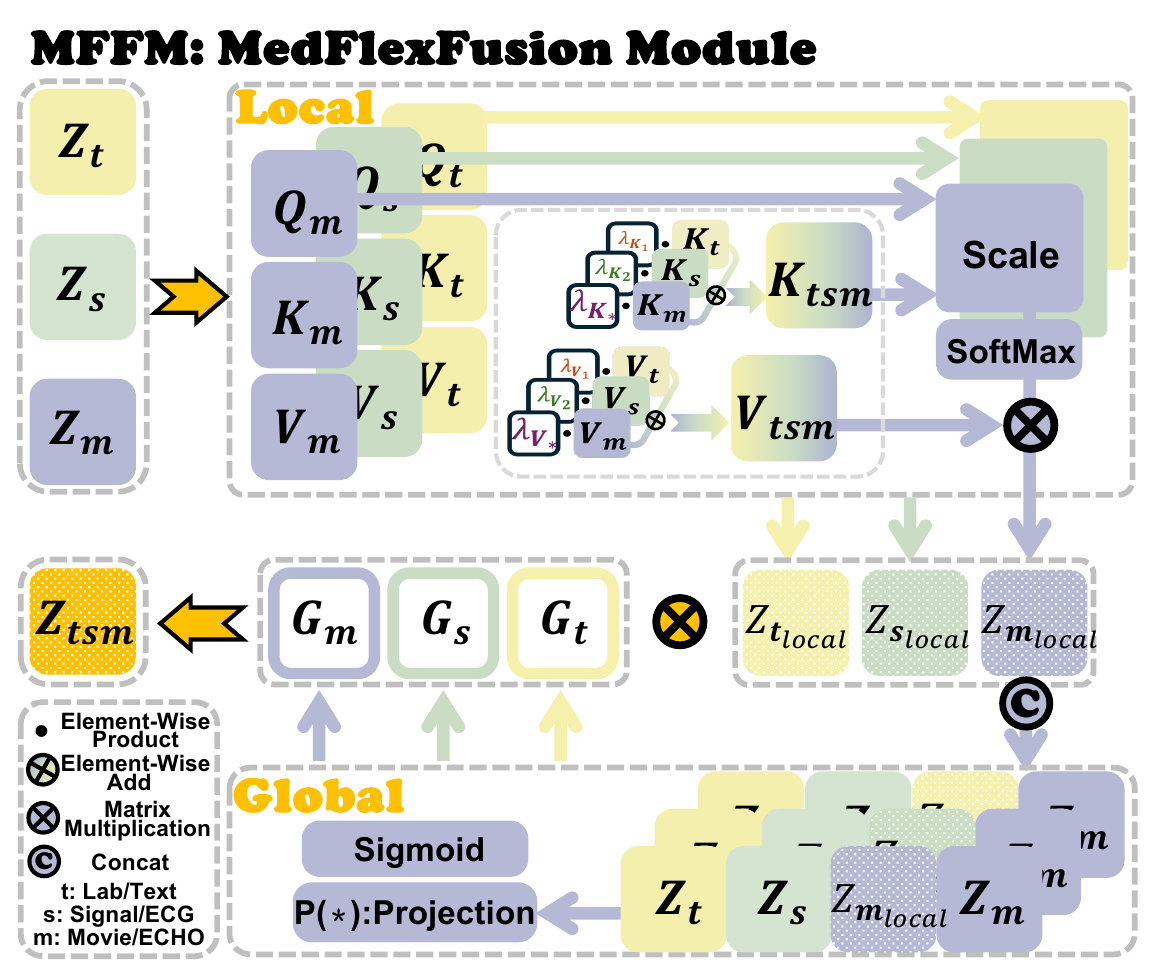}
    \caption{Illustration of the MedFlexFusion Module (MFFM), with local feature fusion (designed for flexible fusion of multiple modalities) and global-context-based gating of fused features.}
    \label{fig:mfm}
\end{figure}

\begin{figure}[t!]
    \centering
    \includegraphics[width=1\linewidth]{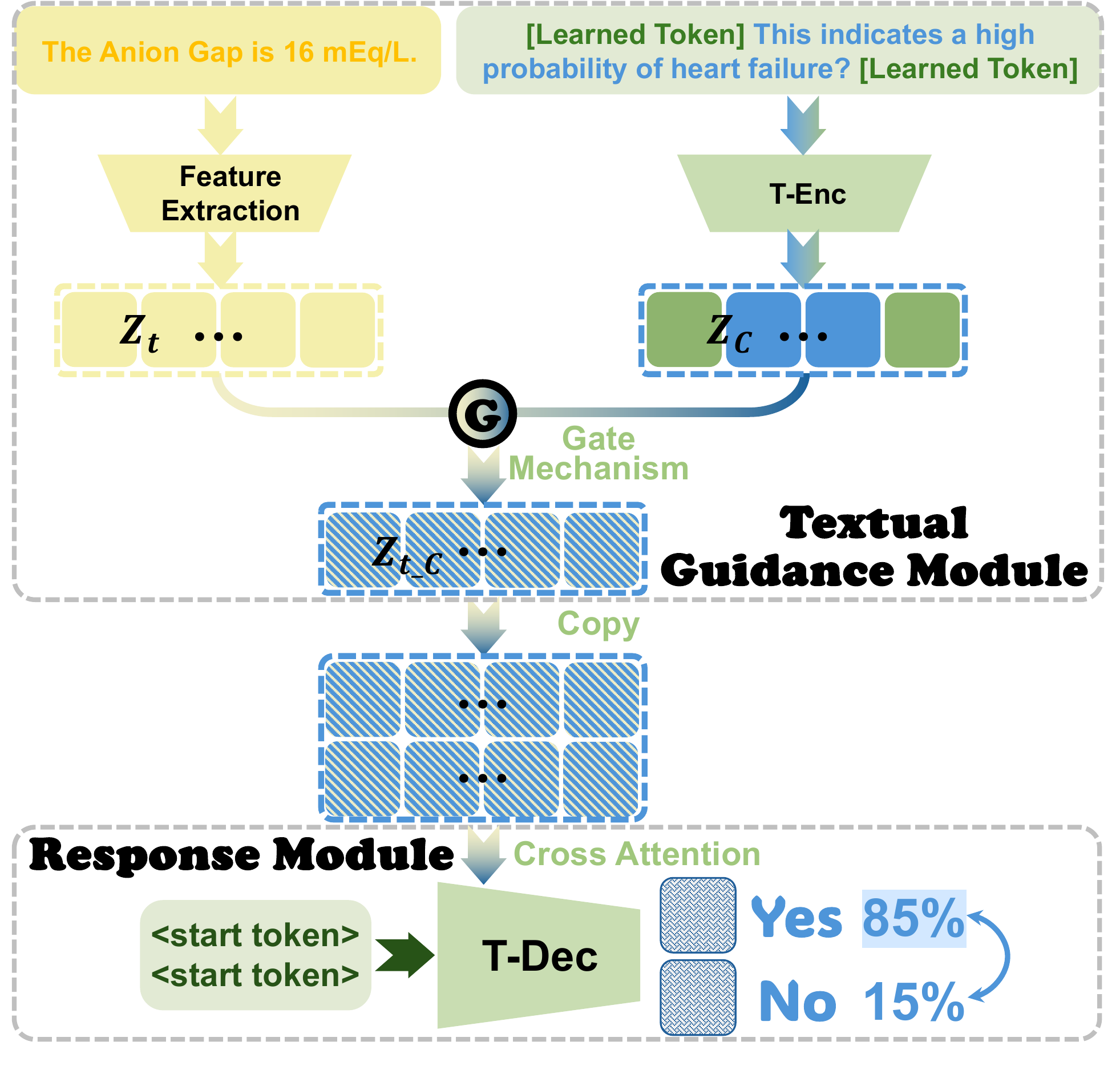}
   \caption{Illustration of textual guidance prediction with a single-modality (text) example: textual guidance module (top) and response module (bottom). For clarity, four vectors and an ellipsis are shown; the actual number of vectors varies with the number of modalities involved.}
    \label{fig:tg}
\end{figure}
\subsection{Problem Statement}
Consider the dataset with $N$ patients:
\[
\Theta_{t,s,m}^i = \{(\Theta_t^i, \Theta_s^i, \Theta_m^i) \mid i = 0, 1, \dots, N-1\},
\]
which consisted of pairs of Labs ($\Theta_t^i$), ECGs ($\Theta_s^i$), and ECHOs ($\Theta_m^i$). Specifically, the Labs were represented in a structured table with $j$ columns:
\[
\Theta_t^i = \{(x_t^{i,0}, x_t^{i,1}, \dots, x_t^{i,j-1})\}.
\]

The ECGs consisted of 12 leads ($L=12$) and they were time series data with $p$ timesteps:
\[
\begin{aligned}
\Theta_s^i =\{ & (x_{s}^{i,0}, x_{s}^{i,1}, \dots, x_{s}^{i,p-1})^0, \\
                & (x_{s}^{i,0}, x_{s}^{i,1}, \dots, x_{s}^{i,p-1})^1, \dots, \\
                & (x_{s}^{i,0}, x_{s}^{i,1}, \dots, x_{s}^{i,p-1})^{L-1} \}.
\end{aligned}
\]

The ECHOs had eleven views ($E=11$), which were provided as sequences of $f$ frames:
\[
\begin{aligned}
\Theta_m^i = \{ & (x_{m}^{i,0}, x_{m}^{i,1}, \dots, x_{m}^{i,f-1})^0, \\
                & (x_{m}^{i,0}, x_{m}^{i,1}, \dots, x_{m}^{i,f-1})^1, \dots, \\
                & (x_{m}^{i,0}, x_{m}^{i,1}, \dots, x_{m}^{i,f-1})^{E-1}\}.
\end{aligned}
\]

\subsection{TGMM Framework}
\subsubsection{Modality-Aware Representation Learning (MARL)}
Under limited availability of large-scale cardiac multimodal datasets, modality-specific representations were extracted via domain-specific foundation models to facilitate modality-aware learning. Specifically, transformer-based Bio-Clinical BERT \cite{alsentzer2019publicly} was employed for the Labs. Natural language text is a more flexible and extensively studied format compared to tabular data, especially when dealing with missing values and context information for numerical data \cite{hegselmann2023tabllm, xie2022unifiedskg}. Inspired by existing works \cite{hegselmann2023tabllm}, a template was employed to transform sentences describing Lab results into a textualized format such as:
\[
\texttt{<column name> of the <object> is <value>}
\] 
This approach outperformed more complex transformation methods, as demonstrated in the comparison in the Appendix A. With the transformed text, the contextualized representations were then extracted using Bio-Clinical BERT as follows: 
\[
Z_t = \{\text{BERT}(\Theta_t^i)\mid i = 0, 1, \dots, N-1 \}.
\]

Then, a transformer-based model, named Spatio-Temporal Masked Electrocardiogram Modelling (ST-MEM)~\cite{naguiding}, was employed to capture spatio-temporal dynamics and the interdependencies across the 12 ECG leads:
\[
Z_s = \{ \text{ST-MEM}(\Theta_s^i) \mid i = 0, 1, \dots, N-1 \}.
\]

For echocardiogram videos, EchoPrime ~\cite{vukadinovic2024echoprime}, a multi-view, video-based foundation model trained on over 12 million video-report pairs, was applied:
\[
Z_m = \{\text{EchoPrime}(\Theta_m^i)\mid i = 0, 1, \dots, N-1 \}.
\]

\subsubsection{MedFlexFusion Module (MFFM)}
MFFM was proposed to extract modality-shared features (see Fig.~\ref{fig:mfm}). It took into account the complementary and similar characteristics of medical data by first extracting local fusion features and then leveraging global context to gate their integration. In addition, this module was designed to support flexible fusion of an arbitrary number of modalities, adapting to available modality combinations.

\medskip
\noindent \textbf{Local.}
Given that information from different medical modalities may exhibit complementarity, similarity, or both, a local fusion mechanism inspired by self- and cross-attention mechanisms was proposed. Firstly, upon the
$
Z_{\text{specific}} = \{ Z_t, Z_s, Z_m \},
$
the Query (Q), Key (K), and Value (V) matrices for different modalities
$
\{(Q_t, K_t, V_t); (Q_s, K_s, V_s); (Q_m, K_m, V_m)\}
$
were derived via \hyperref[eq:query_key_value]{Equation 1}:
\begin{equation}
\begin{pmatrix}
Q \\ 
K \\ 
V
\end{pmatrix}
= Z_{\text{$i\in\{t, s, m\}$}}
\begin{pmatrix}
W_Q \\ 
W_K \\ 
W_V
\end{pmatrix},
\label{eq:query_key_value}
\end{equation}
where $W_Q$, $W_K$, and $W_V$ were the learnable parameters.\\

Then, MFFM was designed to use one modality to query all others, as depicted in \hyperref[eq:local_attention]{Equation 2}.
\begin{equation}
Z_{i_{\text{local}}} = \text{softmax}\left(\frac{Q_i K_{\text{tsm}}^T}{\sqrt{d_{K_{tsm}}}}\right) V_{\text{tsm}}, \label{eq:local_attention}
\end{equation}
where $d_{K_{tsm}}$ was the dimension of  $K_{\text{tsm}}$, $Z_{i_{\text{local}}}$ represented the modality $i$ fused with the local features, $i \in \{t, s, m\}$ corresponded to the index of Labs, ECGs, and ECHOs, respectively. $K_{\text{tsm}}$ and $V_{\text{tsm}}$ were derived from the following equation:
\begin{equation}
\begin{aligned}
\begin{bmatrix}
K_{\text{tsm}} \\ 
V_{\text{tsm}}
\end{bmatrix}
&=
\begin{bmatrix}
\lambda_{k_1} \\ 
\lambda_{v_1}
\end{bmatrix}
\begin{bmatrix}
P(K_t) \\ 
P(V_t)
\end{bmatrix}
+
\begin{bmatrix}
\lambda_{k_2} \\ 
\lambda_{v_2}
\end{bmatrix}
\begin{bmatrix}
P(K_s) \\ 
P(V_s)
\end{bmatrix} \\
&\quad +
\left(1 -
\begin{bmatrix}
\lambda_{k_1} \\ 
\lambda_{v_1}
\end{bmatrix}
-
\begin{bmatrix}
\lambda_{k_2} \\ 
\lambda_{v_2}
\end{bmatrix}
\right)
\begin{bmatrix}
P(K_m) \\ 
P(V_m)
\end{bmatrix},
\end{aligned}
\label{eq:Equation3}
\end{equation}
where $P(*)$ refered to a projection operation that maped features into high-dimensional space, a technique shown to be efficient in various applications \cite{liu2023m}. $\lambda_*$ determined the relative importance of the features from different modalities. These weights enabled selective attention to specific features across modalities (see Equation \ref{eq:weights_4_1} and Equation \ref{eq:weights_4_2}), where the sigmoid function ensured that the weights were constrained within the range of 0 and 1.

\begin{equation}
\begin{bmatrix}
\lambda_{k_1} \\ 
\lambda_{v_1}
\end{bmatrix}
= \text{sigmoid}\left(\frac{P(K_t) + P(K_s) + P(K_m)}{P(V_t) + P(V_s) + P(V_m)}\right),
\label{eq:weights_4_1}
\end{equation}

\begin{equation}
\begin{bmatrix}
\lambda_{k_2} \\ 
\lambda_{v_2}
\end{bmatrix}
= \text{sigmoid}\left(\frac{P(K_t) + P(K_s)}{P(V_t) + P(V_s)}\right).
\label{eq:weights_4_2}
\end{equation}
Note that $\lambda_{k_1}, \lambda_{v_1}$ were associated with three modalities, while $\lambda_{k_2}, \lambda_{v_2}$ were associated with two modalities. All of these parameters were learnable during training, ensuring adaptability across trimodal, bimodal, and even unimodal scenarios, thereby enhancing the generalizability of the model to different modality configurations.

\medskip
\noindent \textbf{Global.} While each modality captured shared features from its own perspective within the context of all modalities, the contribution of each to the final fused representation remained unclear. To address this, a global gating mechanism was proposed to adaptively control modality-wise contributions. Specifically, with the acquired local fused features $\{Z_{t_{\text{local}}}, Z_{s_{\text{local}}}, Z_{m_{\text{local}}}\}$, the gating mechanism ($G_{i}$) was introduced to dynamically control the contribution of these features from a global perspective. Specifically, $G_{i}$ was defined as:
\begin{equation}
G_{i} = \text{sigmoid}\left(P\left(\text{Cat}\left(Z_t, Z_s, Z_{i_{\text{local}}}, Z_m\right)\right)\right)
\label{eq:global_gate}.
\end{equation}

Then, the final fused embedding could be derived as:

\begin{equation}
Z_{tsm} = \sum_{i \in \{t, s, m\}}G_{i} Z_{i_{\text{local}}},
\label{eq:final_fused_embedding}
\end{equation}
where the shared features 
$
Z_{\text{shared}} = \{Z_{ts}, Z_{tm}, Z_{sm}, Z_{tsm}\}
$
were derived by setting $i \in \{t, s\}$, $i \in \{t, m\}$, $i \in \{s, m\}$, and $i \in \{t, s, m\}$, respectively.

Lastly, to further capture complementary and similar features inherent in medical data, both modality-specific and modality-shared features were used, and the final representation could be derived as:
\begin{equation}
Z_{\text{all}} = \text{ReLU}\left(\text{Cat}\left(Z_t, Z_{ts}, Z_s, Z_{sm}, Z_m, Z_{tm}, Z_{tsm}\right)\right)
\label{eq:final_representation},
\end{equation}
where the operator \( \text{Cat}(\cdot) \) denoted the concatenation along the feature dimension, and \( \text{ReLU}(\cdot) \) applied the rectified linear unit activation function to enhance non-linearity and introduce sparsity in the fused feature space.

\subsubsection{Textual Guidance Module (TGM)}
To reduce redundancy, TGM was introduced to extract task-relevant features. Firstly, the semi-soft textual content (SSTC) was defined as consisting of two components: a human-defined segment tailored to the specific task ($C_h$) and a learnable content ($C_l$), as illustrated in Fig.~\ref{fig:tg} (top). The human-defined component ($C_h$) represented task-specific queries that were manually designed with domain knowledge and tokenized into discrete tokens with Bio-ClinicalBERT, while the learnable component ($C_l$) was initialized with tokens from the same tokenizer. This learnable segment enabled the model to adapt to patterns not explicitly defined by humans, thereby ensuring that the proposed SSTC effectively bridged the gap between human comprehension and machine adaptability.

Then, the overall SSTC was formed by concatenating $C_h$ and $C_l$, where $C_h$ could be inserted at any position within $C_l$. This insertion position was treated as a tunable hyperparameter, enabling flexible adaptation for different tasks. This concatenated content was encoded with a single-layer transformer encoder (TE) to derive textual guidance representations.
\[
Z_C = \text{TE}\left(\Theta_{C = \text{Concat}(C_h, C_l)}\right).
\]

Subsequently, the acquired representation was used to dynamically control the contribution of modality-specific and modality-shared features, and to extract task-specific features ($Z_{all\_C}$, see Equation~\ref{eq:9}). Specifically, the gating mechanism ($G_{all}$) in Equation~\ref{eq:global_gate} was employed to dynamically filter features within the contextual information of the task-specific textual content.
\begin{equation}
\begin{aligned}
Z_{\text{all}\_C} = \operatorname{LN} \Bigg( 
    & G_{\text{all}} \odot 
    \left[ 
        \operatorname{softmax} \left( 
            \frac{Q_{\text{all}} K_{\text{all\_C}}^T}{\sqrt{d_{\text{all\_C}}}} 
        \right) 
        V_{\text{all\_C}} 
    \right] \\
    & + \operatorname{DP} \left( Z_{\text{all}}; p_{\text{drop}} \right)
\Bigg),
\end{aligned}
\label{eq:9}.
\end{equation} where $Q_{\text{all}}$ denoted the query of the final representation ($Z_{\text{all}}$), as defined in Equation~\ref{eq:query_key_value}; $K_{\text{all\_C}}$ and $V_{\text{all\_C}}$ were obtained from Equationn~\ref{eq:Equation3} based on both the final representation and the task-specific
textual representations;  $d_{\text{all\_C}}$ was the dimension of  $K_{\text{all\_C}}$; LN refered to layer normalization; DP denoted stochastic depth regularization, parameterized by the drop probability hyperparameter $p_{\text{drop}}$; and $\odot$ denoted element-wise multiplication.

\subsubsection{Response Module}
The response module was architected to facilitate task-specific predictions, leveraging a single-layer Transformer decoder initialized from scratch, as depicted in Fig.~\ref{fig:tg} (bottom). During inference, the extracted task-specific features ($Z_{\text{all}\_C}$) acted as the encoder outputs and were duplicated for each candidate answer, paired with its corresponding SSTC. For each candidate, the decoder processed a shifted target sequence, with a prepended start token. The model then produced outputs for all candidates, with the final prediction being the answer associated with the highest likelihood. Notably, this approach reformulated the classification task as an adversarial comparison, where the model produced candidate outputs for each class and directly contrasts their likelihoods. By leveraging this adversarial mechanism, the model was encouraged to discern subtle differences among candidate responses, thereby enhancing decision robustness.

\subsection{Loss Function}
The proposed TGMM were used for three cardiac tasks: heart disease diagnosis, risk-based prognostic stratification, and information retrieval. Different loss functions were employed for different tasks.
\subsubsection{Diagnosis Task}
The loss function for the disease diagnosis ($L_{\text{dig}}$) consisted of three components:
\begin{equation}
L_{\text{dig}} = \lambda_{\text{lm}} \cdot L_{\text{lm}} + \lambda_{\text{mc}} \cdot L_{\text{mc}} + \lambda_{\text{unlikely}} \cdot L_{\text{unlikely}}
\end{equation}

For $L_{\text{lm}}$ and $L_{\text{mc}}$, they were computed using the cross-entropy loss, defined as
\begin{equation}
L_{\text{CE}} = -\sum_{i=0}^C y_i \cdot \log\left(\hat{y}_i\right),
\end{equation}
where $y_i$ represented the ground truth probability distribution; $\hat{y}_i$ represented the predicted probability for the $ith$ class.

For $L_{\text{unlikely}}$, it was designed to penalize the model for assigning high probabilities to incorrect answer options, thereby encouraging it to prioritize the generation of valid and relevant outputs. The formulation was
\begin{equation}
L_{\text{unlikely}} = -\frac{1}{C - 1} \sum_{j \ne y^*} \log\left(1 - \exp\left(\hat{y}_j\right) + \epsilon\right),
\end{equation}
where \( C \) denoted the total number of candidate answers, \( y^* \) was the index of the correct answer, \( \hat{y}_j \) represented the total log-likelihood of the \( j \)-th candidate answer, and \( \epsilon \) was a small positive constant added for numerical stability to avoid taking the logarithm of zero.

\subsubsection{Prognosis Task}
The diagnosis task focused on the present condition of patients, whereas the prognosis task aimed to predict the likelihood of future disease onset. The loss function was defined as
\begin{equation}
L_{\text{risk}} = \lambda_{\text{dig}} \cdot L_{\text{dig}} + \lambda_r \cdot L_r + \lambda_m \cdot L_m,
\end{equation}
where $L_{\text{dig}}$ was the diagnosis loss, and $\lambda_{\text{dig}}$, $\lambda_r$, and $\lambda_m$ were hyperparameters. 

$L_r$ was the risk loss, which minimized the average negative log partial likelihood of the incident patient set~\cite{katzman2018deepsurv}, and was defined as
\begin{equation}
L_r = -\frac{1}{n_{\delta=1}} \sum_{i:\delta_i=1} \left[\hat{y}_i - \log \sum_{j:t_j \geq t_i} \exp\left(\hat{y}_j\right)\right],
\end{equation}
where $n_{\delta=1}$ was the number of incident patients, and $\hat{y}$ was the predicted risk score.

$L_m$ was the margin ranking loss, which ensured that higher-risk cases receive appropriately larger predicted values, aligning with the ranking constraints of the task. It was defined as:
\begin{equation}
L_m = -\frac{1}{n} \sum_i \max\left(0, -y_i \cdot \left(\hat{y}_i - \hat{y}_j\right) + \text{margin}\right),
\end{equation}
where $n$ was the number of ranked pairs, $\hat{y}_i$ and $\hat{y}_j$ were predicted scores, and the margin parameter enforced a minimum difference between them.

\subsubsection{Information Retrieval}
Contrastive loss was employed for retrieval tasks to minimize the distance between positive pairs and maximize it for negative pairs. The formulation was given by:
\begin{equation}
l_i^{(v \to u)} = -\log \frac{\exp\left(\langle v_i, u_i \rangle / \tau \right)}{\sum_{k=1}^N \exp\left(\langle v_i, u_k \rangle / \tau \right)},
\end{equation}
where \( v_i \) and \( u_i \) denoted the feature embeddings of the \( i \)-th sample from two different modalities (e.g., modality \( v \) and modality \( u \)), $\langle v_i, u_i \rangle$ represented the cosine similarity, and $\tau \in \mathbb{R}^+$ represented a temperature parameter.

\section{Experiments}
\subsection{Experimental Settings}
\noindent \textbf{Implementation details.} For laboratory test data, the Bio-ClinicalBERT tokenizer was used to tokenize text inputs to 512 tokens. Electrocardiograms recordings were preprocessed by resampling to 250 Hz, applying a 0.67 – 40 Hz bandpass filter, and removing noise and baseline drift~\cite{naguiding}. For echocardiograms, the input video size was set to $11 \times 224 \times 224 \times 16 \times 3$, with a stride of 2 and 16 frames to enhance temporal context. All modalities were represented as 768-dimensional vectors. 

The model was trained with the Adafactor optimizer, a ReduceLROnPlateau scheduler, and mixed precision. The dataset was split with iterative stratification at ratio of 5:1:1 ~\cite{szymanski2017network}. For risk stratification, a two-fold stratified cross-validation approach was applied~\cite{chen2025large}, with 20\% of the training data allocated for validation. This procedure was repeated five times to ensure a robust evaluation. Further details are provided in the Appendix B. 

\medskip
\noindent \textbf{Evaluation metrics.} The diagnosis task was evaluated using accuracy-based metrics, including the area under the receiver operating characteristic curve (AUC) and overall accuracy~\cite{yu2021evaluation}, with 95\% confidence intervals reported. Risk stratification was assessed using the concordance index (C-index)~\cite{harrell1982evaluating}. For the retrieval task, performance was evaluated with the label ranking average precision (LRAP), Recall@1 and Recall@2 for identifying the correct candidate.

\medskip
\noindent \textbf{Baseline models.} The proposed framework was evaluated against contrastive learning–based models such as EchoPrime~\cite{vukadinovic2024echoprime}, which first perform feature alignment and subsequently apply an additional classifier for downstream tasks. Additionally, its performance was assessed in comparison to advanced larger models, including Qwen2.5-VL from Alibaba~\cite{wang2024qwen2} and Janus-Pro from Deepseek AI~\cite{chen2025janus}. Finally, the proposed fusion techniques were evaluated against existing multi-fusion approaches, including  CMTA~\cite{zhou2023cross}, DeepAVFusion~\cite{mo2024unveiling}, Text-IF~\cite{yi2024text}, MultiSurv~\cite{vale2021long}, and a traditional mean-concatenation (MeanCat) method~\cite{steyaert2023multimodal}. Note that all compared approaches employed the same foundation models to ensure a fair comparison.

\begin{table*}[htbp]
\centering
\caption{HF Performance Comparison across Modalities and Models}
\vspace{-8pt}
\small
\begin{adjustbox}{width=\textwidth}
\begin{tabular}{cccccccccc}
\toprule
\multirow{3}{*}{\textbf{\begin{tabular}[c]{@{}c@{}}Task\end{tabular}}} & 
\multirow{3}{*}{\textbf{\begin{tabular}[c]{@{}c@{}}Metric\end{tabular}}} & 
\multirow{3}{*}{\textbf{\begin{tabular}[c]{@{}c@{}}Model\end{tabular}}} & 
\multicolumn{7}{c}{\textbf{Modalities}} \\
\cmidrule{4-10}
& & & \textbf{Lab} & \textbf{ECG} & \textbf{ECHO} & \textbf{Lab+ECG} & \textbf{Lab+ECHO} & \textbf{ECG+ECHO} & \makecell{\textbf{Lab+} \\ \textbf{ECG+ECHO}} \\
\midrule

\multirow{8}{*}{\textbf{\begin{tabular}[c]{@{}c@{}}Diagnosis\end{tabular}}} & \multirow{4}{*}{\textbf{ACC}}  
& EchoPrime~\cite{vukadinovic2024echoprime} & — & — & — & 0.65 (0.60, 0.70) & 0.74 (0.69, 0.79) & 0.73 (0.68, 0.78) & — \\
& & Qwen2.5-VL~\cite{wang2024qwen2} & 0.72(0.67, 0.77)* & 0.71 (0.66, 0.76) & 0.77 (0.72, 0.82) & 0.72 (0.67, 0.77) & 0.77 (0.73, 0.82) & \underline{0.79 (0.74, 0.83)} & 0.78 (0.73, 0.83) \\
& & Janus-Pro~\cite{chen2025janus} & 0.74 (0.69, 0.79) & 0.73 (0.68, 0.78) & 0.76 (0.71, 0.81) & 0.80(0.75, 0.84)* & 0.83 (0.78, 0.87) & 0.78 (0.73, 0.83) & 0.84 (0.79, 0.88) \\
& & TGMM & \underline{0.77 (0.72, 0.82)} & \underline{0.76 (0.71, 0.81)} & \underline{0.79 (0.74,0.83)} & \underline{0.83 (0.78, 0.87)} & \underline{0.84 (0.80, 0.88)} & 0.78 (0.73, 0.82) & \underline{0.86 (0.81, 0.89)} \\
\cmidrule{2-10}

& \multirow{4}{*}{\textbf{AUC}}
& EchoPrime~\cite{vukadinovic2024echoprime} & — & — & — & 0.71 (0.65, 0.77) & 0.82 (0.77, 0.87) & 0.82 (0.77, 0.86) & — \\
& & Qwen2.5-VL~\cite{wang2024qwen2} & 0.76(0.71, 0.82)* & 0.79 (0.73, 0.84) & \underline{0.85 (0.80, 0.89)} & 0.78 (0.72, 0.83) & 0.83 (0.79, 0.88) & 0.83 (0.79, 0.88) & 0.85 (0.81, 0.90) \\
& & Janus-Pro~\cite{chen2025janus} & 0.76 (0.70, 0.81) & 0.80 (0.75, 0.84) & 0.84 (0.80, 0.89) & 0.86(0.82, 0.90)* & \underline{0.90 (0.87, 0.94)} & \underline{0.86 (0.82, 0.90)} & \underline{0.91 (0.87, 0.94)} \\
& & TGMM & \underline{0.82 (0.77, 0.86)} & \underline{0.83 (0.78, 0.87)} & \underline{0.85 (0.80, 0.89)} & \underline{0.89 (0.85, 0.92)} & \underline{0.90 (0.87, 0.94)} & 0.85 (0.81, 0.89) & \underline{0.91 (0.87, 0.94)} \\
\midrule

\multirow{4}{*}{\textbf{\begin{tabular}[c]{@{}c@{}}Prognosis\end{tabular}}} & \multirow{4}{*}{\textbf{C-index}} 
& EchoPrime~\cite{vukadinovic2024echoprime} & — & — & — & 0.54±0.05 & 0.56±0.04 & 0.53±0.04 & — \\
& & Qwen2.5-VL~\cite{wang2024qwen2} & 0.51±0.05 & 0.49±0.04 & 0.51±0.04 & 0.51±0.03 & 0.51±0.04 & 0.53±0.05 & 0.52±0.03 \\
& & Janus-Pro~\cite{chen2025janus} & 0.52±0.06 & 0.51±0.03 & 0.54±0.03 & 0.50±0.03 & 0.53±0.03 & 0.51±0.05 & 0.54±0.04 \\
& & TGMM & \underline{0.53±0.04} & \underline{0.55±0.03} & \underline{0.58±0.04} & \underline{0.57±0.04} & \underline{0.59±0.03} & \underline{0.59±0.04} & \underline{0.61±0.05} \\
\bottomrule
\multicolumn{10}{l}{\scriptsize{Note: — indicates modality not supported by the model; Underlined values indicate best performance within each modality; Values in parentheses represent 95\% confidence intervals; C-index values shown with standard deviation;}} \\[-4pt]
\multicolumn{10}{l}{\scriptsize{\phantom{Note: }$^*$No significant difference compared to the proposed method under the same modality setting (p$>$0.05); ACC = accuracy; TGMM = Textual Guidance Multimodal fusion framework for Multiple Tasks; HF = Heart Failure.}} \\
\end{tabular}
\label{tab:task12}
\end{adjustbox}
\end{table*}
\begin{table*}[htbp]
\centering
\caption{Performance Metrics for Cross-Modal Medical Data Retrieval}
\vspace{-8pt}
\small
\begin{adjustbox}{width=\textwidth}
\begin{tabular}{ccccccc}
\toprule
\multirow{2}{*}{\textbf{Metric}} & \multicolumn{6}{c}{\textbf{Retrieval Direction}} \\
\cmidrule{2-7}
& \textbf{Lab→ECG} & \textbf{ECG→Lab} & \textbf{Lab→ECHO} & \textbf{ECHO→Lab} & \textbf{ECG→ECHO} & \textbf{ECHO→ECG} \\
\midrule
\textbf{LRAP} & 0.61 (0.58, 0.64) & 0.63 (0.59, 0.66) & 0.74 (0.71, 0.77) & 0.73 (0.70, 0.76) & 0.66 (0.63, 0.70) & 0.65 (0.62, 0.67) \\
\midrule
\textbf{Recall@1} & 36.04\% (31.17, 40.58) & 37.99\% (33.43, 42.86) & 54.55\% (50.00, 59.42) & 54.87\% (50.00, 59.42) & 43.18\% (38.64, 48.05) & 40.91\% (36.36, 45.78) \\
\midrule
\textbf{Recall@2} & 67.53\% (62.99, 72.08) & 68.83\% (63.64, 74.03) & 81.49\% (77.59, 85.39) & 78.90\% (74.34, 83.44) & 70.45\% (65.91, 75.65) & 71.43\% (67.21, 75.97) \\
\bottomrule
\multicolumn{7}{l}{\scriptsize{Note: Values in parentheses represent 95\% confidence intervals; LRAP = Label Ranking Average Precision}} \\
\end{tabular}
\label{tab:task3}
\end{adjustbox}
\end{table*}
\subsection*{5.2. Results and Analysis}
\begin{figure}[t!]
    \centering
    \includegraphics[width=1\linewidth]{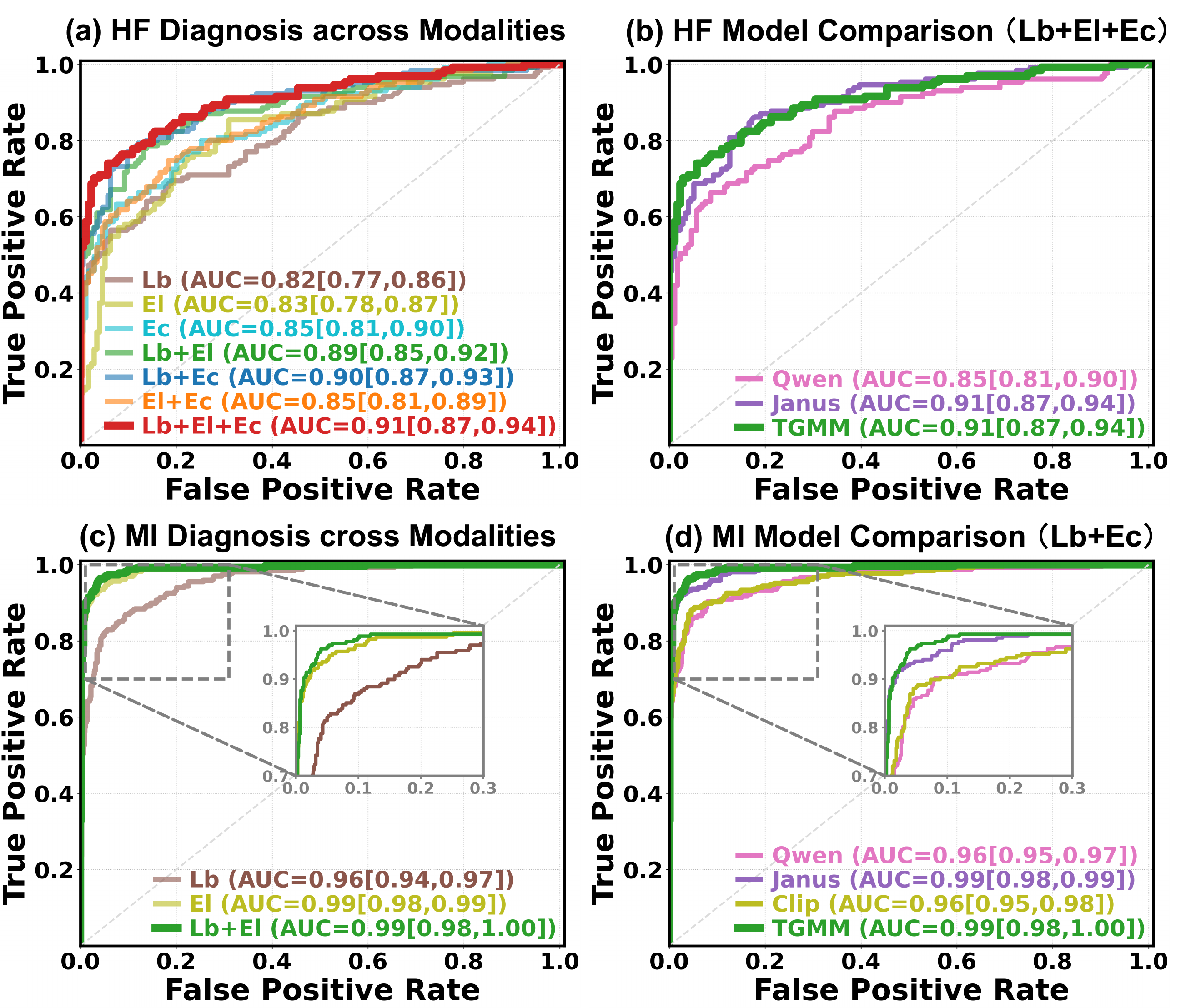}
    \caption{ROC curve with x-axis as False Positive Rate and y-axis as True Positive Rate. (a, b) Heart failure (HF) diagnosis; (c, d) Myocardial infarction (MI) diagnosis. (a, c) Comparison of uni-, bi-, and trimodal approaches: Lb (laboratory test results), El (electrocardiograms), Ec (echocardiograms). (b, d) Performance comparison between the proposed model and baseline models.}
    \label{fig:auroc}
\end{figure}

Table~\ref{tab:task12} presented the performance of the proposed model across different modality settings: unimodal (Labs, ECGs, or ECHOs), bimodal (combinations of two modalities), and trimodal (all three modalities), with corresponding AUROC curves shown in Figure~\ref{fig:auroc}. Among the unimodal settings, Labs exhibited the lowest performance, with an AUC of 0.82 (95\% CI: 0.77–0.86) for diagnosis and a C-index of 0.53±0.04 for risk stratification. The combination of ECGs and ECHOs yielded the lowest bimodal performance for diagnosis (0.85, 95\% CI: 0.81–0.89), while the combination of Labs and ECGs showed the lowest performance for risk stratification (0.57±0.04). The trimodal approach achieved the highest performance, with an AUC of 0.91 (95\% CI: 0.87–0.94) for diagnosis and 0.61±0.05 for risk stratification. Wilcoxon signed-rank tests confirmed significant differences in performance between modality combinations (p $<$ 0.05).

\begin{figure}[t!]
    \centering
    \includegraphics[width=1\linewidth]{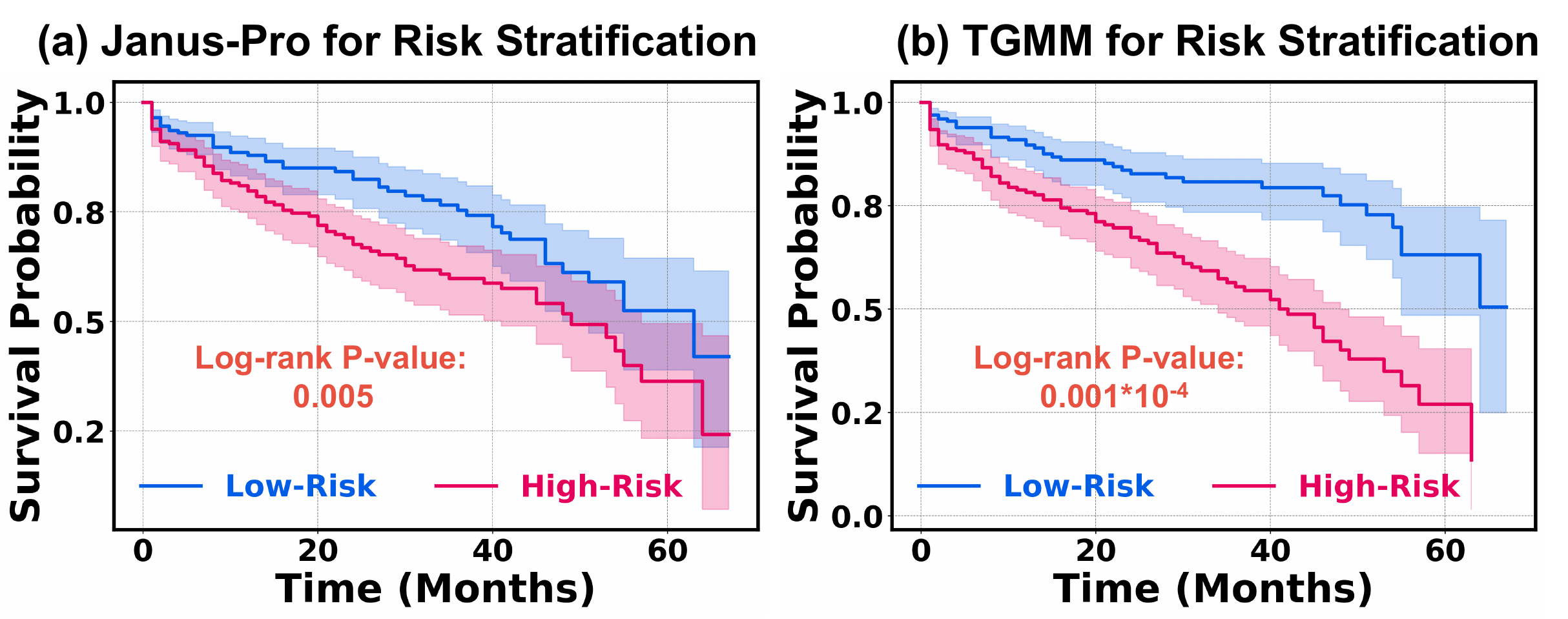}
    \caption{Kaplan-Meier Curves for trimodal Risk Stratification: (a) Janus-Pro, (b) Proposed Textual Guidance Multimodal fusion framework for Multiple Tasks (TGMM).}
    \label{fig:risk}
\end{figure}

Results for the three different clinical tasks are presented in Tables~\ref{tab:task12}, ~\ref{tab:task3} and Figures~\ref{fig:auroc}(a-b). Specifically, in the HF diagnosis task, the model achieved its highest performance with an AUC of 0.91 (95\% CI: 0.87–0.94) and an accuracy of 0.86 (95\% CI: 0.81--0.89). For risk stratification, it attained a maximum C-index of 0.61±0.05, and the Kaplan–Meier (KM) survival curves, as shown in Figure~\ref{fig:risk}, illustrated the differences in survival probabilities between risk groups. Lastly, the proposed framework showed its capability in information retrieval (see Table~\ref{tab:task3}), achieving a LRAP of 0.74 (95\% CI: 0.71--0.77) when using laboratory information to retrieve the echocardiogram.

\medskip
\noindent \textbf{Comparison results.} The results in Table~\ref{tab:task12}, Figure~\ref{fig:auroc}, and Figure~\ref{fig:risk} indicated that the proposed method consistently outperformed other models in HF diagnosis across most modality combinations, and achieved a higher C-index in risk prediction across all modality combinations. In the ECG+ECHO fusion setting, Qwen2.5-VL achieved higher accuracy (0.79 vs. 0.78 in TGMM, \( p < 0.05 \)) and Janus-Pro yielded a better AUC (0.86 vs. 0.85 in TGMM, \( p < 0.05 \)) for HF diagnosis, but the proposed new method showed better performance in risk stratification (0.59 vs. 0.53 in Qwen2.5-VL, vs. 0.51 in Janus-Pro, $p<0.05$). For the trimodal setting, the proposed approach achieved a comparable AUC but significantly higher accuracy than the larger Janus-Pro in diagnosis, and a higher C-index in risk prediction (0.61 vs. 0.54; $p<0.05$). Meanwhile, KM curves presented in Figure~\ref{fig:risk} showed that the proposed trimodal model produced narrower CIs than Janus-Pro. Note that EchoPrime lacked the ability for trimodal fusion.

\medskip
\noindent\textbf{Model performance on MI datasets.} To show the broader applicability of the proposed framework, another public dataset was used for MI diagnosis, which consists of patient information and ECGs (7,172 Normal and 2,975 MI samples, detailed in Appendix A.2). The results in Table~\ref{tab:MI} showed that the fusion of these two modalities achieved an accuracy of 0.96, which was comparable to using ECGs alone, but significantly higher than using Labs alone (0.91, \( p < 0.05 \)). Compared to other models, the proposed method achieved performance comparable to the larger Janus-Pro, while outperforming EchoPrime (0.91, \( p < 0.05 \)) and Qwen2.5-VL (0.93, \( p < 0.05 \)). The corresponding AUROC curves were shown in Figure~\ref{fig:auroc}(c,d). 
\begin{table*}[htbp]
\centering
\caption{MI Performance Comparison across Modalities and Models}
\vspace{-8pt}
\scriptsize
\setlength{\tabcolsep}{2pt}  
\begin{tabular}{ccccc}
\toprule
\textbf{Metric} & \textbf{Model} & \textbf{Lab} & \textbf{ECG} & \textbf{Lab+ECG} \\
\midrule

\multirow{4}{*}{\textbf{ACC}} 
& EchoPrime~\cite{vukadinovic2024echoprime} & — & — & 0.91 (0.88, 0.93)\\
& Qwen2.5-VL~\cite{wang2024qwen2} & 0.73 (0.70, 0.76) & 0.93 (0.91, 0.95) & 0.93 (0.91, 0.94)\\
& Janus-Pro~\cite{chen2025janus} & 0.88 (0.86, 0.90) & 0.93 (0.91, 0.94) & \underline{0.96 (0.95, 0.97)}\\
& TGMM & \underline{0.91 (0.89, 0.93)} & \underline{0.96 (0.95, 0.97)} & \underline{0.96 (0.95, 0.97)}\\
\midrule

\multirow{4}{*}{\textbf{AUC}} 
& EchoPrime~\cite{vukadinovic2024echoprime} & — & — & 0.96 (0.95, 0.98)\\
& Qwen2.5-VL~\cite{wang2024qwen2} & 0.63 (0.59, 0.67) & 0.97 (0.96, 0.98) & 0.96 (0.95, 0.97)\\
& Janus-Pro~\cite{chen2025janus} & 0.95 (0.94, 0.96) & 0.97 (0.95, 0.98) & \underline{0.99 (0.98, 0.99)}\\
& TGMM & \underline{0.96 (0.94, 0.97)} & \underline{0.99 (0.98, 1.00)} & \underline{0.99 (0.98, 1.00)}\\
\bottomrule
\multicolumn{5}{l}{\tiny{Note: ACC = accuracy; MI = Myocardial infarction; TGMM = Textual Guidance Multimodal fusion framework for Multiple Tasks;}} \\[-3pt]
\multicolumn{5}{l}{\tiny{\phantom{Note: }— indicates modality not supported by the model;}} \\[-3pt]
\multicolumn{5}{l}{\tiny{\phantom{Note: }Values in parentheses represent 95\% confidence intervals;}} \\[-3pt]
\multicolumn{5}{l}{\tiny{\phantom{Note: }Underlined values indicate best performance within each modality;}}\\[-3pt]
\multicolumn{5}{l}{\tiny{\phantom{Note: }All models differ significantly under the same modality setting (p$<$0.05).}} \\
\end{tabular}
\label{tab:MI}
\end{table*}


\subsection*{5.3. Ablation Studies}
\noindent\textbf{MedFlexFusion module (MFFM).}  
Table~\ref{tab:cp} compared the proposed MFFM with baseline fusion strategies. While most fusion methods could not perform three-modality fusion, the proposed model successfully integrated all three modalities, and surpassing the next-best model in HF accuracy (0.86 vs. 0.84 in Meancat, \( p < 0.05 \) ). For risk stratification, it attained a C-index of 0.61, outperforming the next-best model with a C-index of 0.54 (\( p < 0.05 \)).  For the bimodal setting, the proposed method achieved good performance for both tasks. Although Meancat showed comparable results in HF diagnosis with Labs and ECHOs fusion (p=0.1), the proposed method substantially outperformed it in risk assessment metrics (0.59 vs. 0.53, $p < 0.05$), highlighting the balanced effectiveness of the proposed approach across different clinical tasks. This balance was further supported by comparisons with DeepAVFusion, which, despite not supporting trimodal fusion, performed comparably in risk assessment, while the proposed method consistently outperformed it in diagnosis across various bimodal combinations.

\begin{table*}[htbp]
\centering
\caption{Comparative Evaluation of Multi-modal Fusion Methods}
\vspace{-8pt}
\small
\begin{adjustbox}{width=\textwidth}
\begin{tabular}{ccccccc}
\toprule
\multirow{2}{*}{\textbf{Task}} & \multirow{2}{*}{\textbf{Metric}} & \multirow{2}{*}{\textbf{Model}} & \multicolumn{4}{c}{\textbf{Modalities}} \\
\cmidrule{4-7}
& & & \textbf{Lab+ECG} & \textbf{Lab+ECHO} & \textbf{ECG+ECHO} & \textbf{Lab+ECG+ECHO} \\
\midrule

\multirow{12}{*}{\textbf{Diagnosis}} & \multirow{6}{*}{\textbf{ACC}} 
& CMTA~\cite{zhou2023cross} & 0.73 (0.68, 0.78) & 0.68 (0.63, 0.73) & 0.61 (0.55, 0.66) & — \\
& & DeepAVFusion~\cite{mo2024unveiling} & 0.79 (0.75, 0.84) & 0.82 (0.78, 0.87) & 0.70 (0.66, 0.76) & — \\
& & Text-IF~\cite{yi2024text} & 0.57 (0.51, 0.63) & 0.57 (0.51, 0.63) & 0.57 (0.51, 0.63) & — \\
& & MultiSurv~\cite{vale2021long} & 0.73 (0.69, 0.78) & 0.78 (0.73, 0.83) & 0.76 (0.71, 0.81) & 0.78 (0.73, 0.83) \\
& & Meancat~\cite{steyaert2023multimodal} & 0.77 (0.72, 0.82) & 0.83(0.79,0.87)*& 0.77 (0.72, 0.81) & 0.84 (0.79, 0.88) \\
& & TGMM & \underline{0.83 (0.78, 0.87)} & \underline{0.84 (0.80, 0.88)} & \underline{0.78 (0.73, 0.82)} & \underline{0.86 (0.81, 0.89)} \\
\cmidrule{2-7}

& \multirow{6}{*}{\textbf{AUC}} 
& CMTA~\cite{zhou2023cross} & 0.75 (0.69, 0.80) & 0.73 (0.67, 0.78) & 0.60 (0.53, 0.66) & — \\
& & DeepAVFusion~\cite{mo2024unveiling} & 0.86 (0.81, 0.90) & 0.89 (0.85, 0.92) & 0.85 (0.81, 0.89) & — \\
& & Text-IF~\cite{yi2024text} & 0.53 (0.46, 0.59) & 0.62 (0.55, 0.68) & 0.62 (0.55, 0.68) & — \\
& & MultiSurv~\cite{vale2021long} & 0.86 (0.82, 0.90) & 0.86 (0.82, 0.90) & 0.84 (0.80, 0.89) & 0.88 (0.84, 0.91) \\
& & Meancat~\cite{steyaert2023multimodal} & 0.88 (0.84, 0.92) & 0.89(0.85,0.93)*& 0.85 (0.80, 0.89) & 0.90 (0.87, 0.93) \\
& & TGMM & \underline{0.89 (0.84, 0.92)} & \underline{0.90 (0.87, 0.93)} & \underline{0.85 (0.81, 0.89)} & \underline{0.91 (0.87, 0.94)} \\
\midrule

\multirow{6}{*}{\textbf{Prognosis}} & \multirow{6}{*}{\textbf{C-index}} 
& CMTA~\cite{zhou2023cross} & 0.51 ± 0.04 & 0.51 ± 0.04 & 0.52 ± 0.04 & — \\
& & DeepAVFusion~\cite{mo2024unveiling} & \underline{0.57 ± 0.04} & 0.58±0.04* & \underline{0.59 ± 0.05} & — \\
& & Text-IF~\cite{yi2024text} & 0.52 ± 0.04 & 0.52 ± 0.04 & 0.53 ± 0.03 & — \\
& & MultiSurv~\cite{vale2021long} & 0.51 ± 0.06 & 0.56 ± 0.04 & 0.53 ± 0.04 & 0.53 ± 0.04 \\
& & Meancat~\cite{steyaert2023multimodal} & 0.52 ± 0.04 & 0.53 ± 0.03 & 0.53 ± 0.04 & 0.54 ± 0.03 \\
& & TGMM & \underline{0.57 ± 0.04} & \underline{0.59 ± 0.03} & \underline{0.59 ± 0.04} & \underline{0.61 ± 0.05} \\
\bottomrule
\multicolumn{7}{l}{\scriptsize{Note: — indicates modality not supported by the model; Underlined values indicate best performance; Values in parentheses represent 95\% confidence intervals;}} \\[-4pt]
\multicolumn{7}{l}{\scriptsize{\phantom{Note: }C-index values shown with standard deviation; ACC = accuracy; TGMM = Textual Guidance Multimodal fusion framework for Multiple Tasks;}}\\[-4pt]
\multicolumn{7}{l}{\scriptsize{\phantom{Note: }$^*$No significant difference compared to the proposed method under the same modality setting (p$>$0.05)}} \\
\end{tabular}
\label{tab:cp}
\end{adjustbox}
\end{table*}
\begin{table*}[htbp]
\centering
\caption{Ablation Study of Different Model Components across Modalities}
\vspace{-8pt}
\small
\begin{adjustbox}{width=\textwidth}
\begin{tabular}{ccccccccccccc}
\toprule
\multirow{2}{*}{\textbf{Task}} & \multirow{2}{*}{\textbf{Metric}} & \multicolumn{4}{c}{\textbf{Component Used}} & \multicolumn{7}{c}{\textbf{Modalities}} \\
\cmidrule(lr){3-6} \cmidrule(lr){7-13}
& & \textbf{Human} & \textbf{Learned}& \textbf{Response}& \textbf{Linear} & \textbf{Lab} & \textbf{ECG} & \textbf{ECHO} & \textbf{Lab+ECG} & \textbf{Lab+ECHO} & \textbf{ECG+ECHO} & \makecell{\textbf{Lab+} \\ \textbf{ECG+ECHO}} \\
\midrule

\multirow{8}{*}{\textbf{Diagnosis}} & \multirow{4}{*}{\textbf{ACC}} 
& & & \checkmark& & \underline{0.77(0.72,0.82)}& 0.74(0.70,0.79)& 0.78(0.73,0.83)& 0.79(0.75,0.84)& 0.79(0.74,0.83)& 0.75(0.70,0.80)& 0.81(0.76,0.85)\\
& & & & & \checkmark& 0.75(0.70,0.80)& 0.70(0.65,0.75)& 0.73(0.68,0.78)& 0.76(0.72,0.81)& 0.75(0.70,0.80)& 0.71(0.66,0.76)
& 0.75(0.71,0.80)\\
& & \checkmark & & \checkmark& & 0.75(0.70,0.80)
& \underline{0.76(0.71,0.80)}
& \underline{0.79(0.74,0.84)}
& 0.82(0.78,0.86)
& \underline{0.84(0.79,0.88)}
& 0.76(0.71,0.81)
& 0.82(0.77,0.86)\\
& & \checkmark & \checkmark & \checkmark &  & \underline{0.77(0.72,0.82)}
& \underline{0.76(0.71,0.81)}
& \underline{0.79(0.74,0.83)}& \underline{0.83(0.78,0.87)}
& \underline{0.84(0.80,0.88)}
& \underline{0.78(0.73,0.82)}
& \underline{0.86(0.81,0.89)}\\
\cmidrule(lr){2-13}

& \multirow{4}{*}{\textbf{AUC}} 
& & & \checkmark& & 0.80(0.74,0.84)
& 0.81(0.76,0.86)
& 0.84(0.79,0.88)
& 0.87(0.82,0.91)
& 0.85(0.80,0.89)
& 0.82(0.76,0.86)
& 0.90(0.87,0.93)\\
& & & & & \checkmark& 0.75(0.69,0.81)
& 0.70(0.63,0.76)
& 0.71(0.65,0.77)
& 0.75(0.69,0.81)
& 0.81(0.75,0.86)
& 0.78(0.73,0.83)
& 0.81(0.76,0.86)\\
& & \checkmark & & \checkmark& & \underline{0.82(0.77,0.87)}
& 0.81(0.76,0.86)
& \underline{0.85(0.81,0.89)}
& 0.87(0.83,0.91)
& 0.89(0.85,0.93)
& 0.84(0.79,0.88)
& 0.89(0.85,0.93)\\
& & \checkmark & \checkmark & \checkmark &  & \underline{0.82(0.77,0.86)}
& \underline{0.83(0.78,0.87)}
& \underline{0.85(0.81,0.89)}& \underline{0.89(0.85,0.92)}
& \underline{0.90(0.87,0.94)}
& \underline{0.85(0.81,0.89)}
& \underline{0.91(0.87,0.94)}\\
\midrule

\multirow{4}{*}{\textbf{Prognosis}} & \multirow{4}{*}{\textbf{C-index}} 
& & & \checkmark& & \underline{0.53±0.04}
& \underline{0.55±0.03}
& \underline{0.58±0.05}
& 0.51±0.04
& 0.57±0.03
& 0.55±0.04
& 0.57±0.05\\
& & & & & \checkmark& 0.51±0.04
& 0.50±0.05
& 0.50±0.06
& 0.52±0.03
& 0.50±0.05
& 0.52±0.03
& 0.51±0.05\\
& & \checkmark & & \checkmark& & \underline{0.53±0.03}
& 0.54±0.04
& \underline{0.58±0.02}& 0.54±0.02
& 0.58±0.03
& 0.53±0.02
& 0.57±0.05\\
& & \checkmark & \checkmark & \checkmark &  & \underline{0.53±0.04}
& \underline{0.55±0.03}
& \underline{0.58±0.04}
& \underline{0.57±0.04}
& \underline{0.59±0.03}
& \underline{0.59±0.04}
& \underline{0.61±0.05}\\

\bottomrule
\multicolumn{13}{l}{\scriptsize{Note: Underlined values indicate best performance; Values in parentheses represent 95\% confidence intervals; C-index values shown with standard deviation;}} \\[-4pt]
\multicolumn{13}{l}{\scriptsize{\phantom{Note: } ACC = accuracy; TGMM = Textual Guidance Multimodal fusion framework for Multiple Tasks; All results differ significantly under the same modality setting (p$<$0.05).}} \\
\end{tabular}
\label{tab:ab}
\end{adjustbox}
\end{table*}

\medskip
\noindent\textbf{Textual guidance module (TGM).} Table~\ref{tab:ab} presents the performance of the framework under three conditions: without the TGM, with only human-defined textual guidance, and with the proposed combination of human-defined and learned textual guidance. The results indicated that relying solely on human-defined textual guidance could sometimes lead to performance degradation. Thus, for example, compared to the absence of textual guidance, for instance, using Labs for HF diagnosis resulted in lower accuracy when guided solely by human-defined text (0.75 vs 0.77, p$<$0.05). Similarly, for risk prediction with ECGs and ECHOs, the C-index decreased from 0.55 to 0.53 (p$<$0.05). In contrast, the proposed hybrid approach, which integrates learned and human-defined textual content, consistently prevented such degradation. It is noteworthy that in risk stratification, the unimodal approach exhibited limited benefits when combined with the TGM, whereas the bimodal and trimodal configurations yielded notable gains, with the highest improvement reaching 4\% (e.g., in ECG-ECHO and in trimodal).

\medskip
\noindent\textbf{Response module.} Table~\ref{tab:ab} also compares the proposed response module with a standard linear prober for downstream tasks. When the response module was replaced with the linear prober, performance decreased across different modality fusion strategies. In the trimodal setting, for example, accuracy performance dropped from 0.81 to 0.75 in HF diagnosis, and from a C-index of 0.57 to 0.51 in risk stratification (p $<$ 0.05). To determine whether the superior performance was attributed to the compatibility of the module with the proposed framework or the intrinsic effectiveness of the module itself, the downstream component in other modality fusion strategies was substituted with the proposed response module. The results showed performance improvements in most cases, as detailed in Appendix C, which also includes additional implementation details of the linear prober.

\begin{figure}[t!]
    \centering
    \includegraphics[width=1\linewidth]{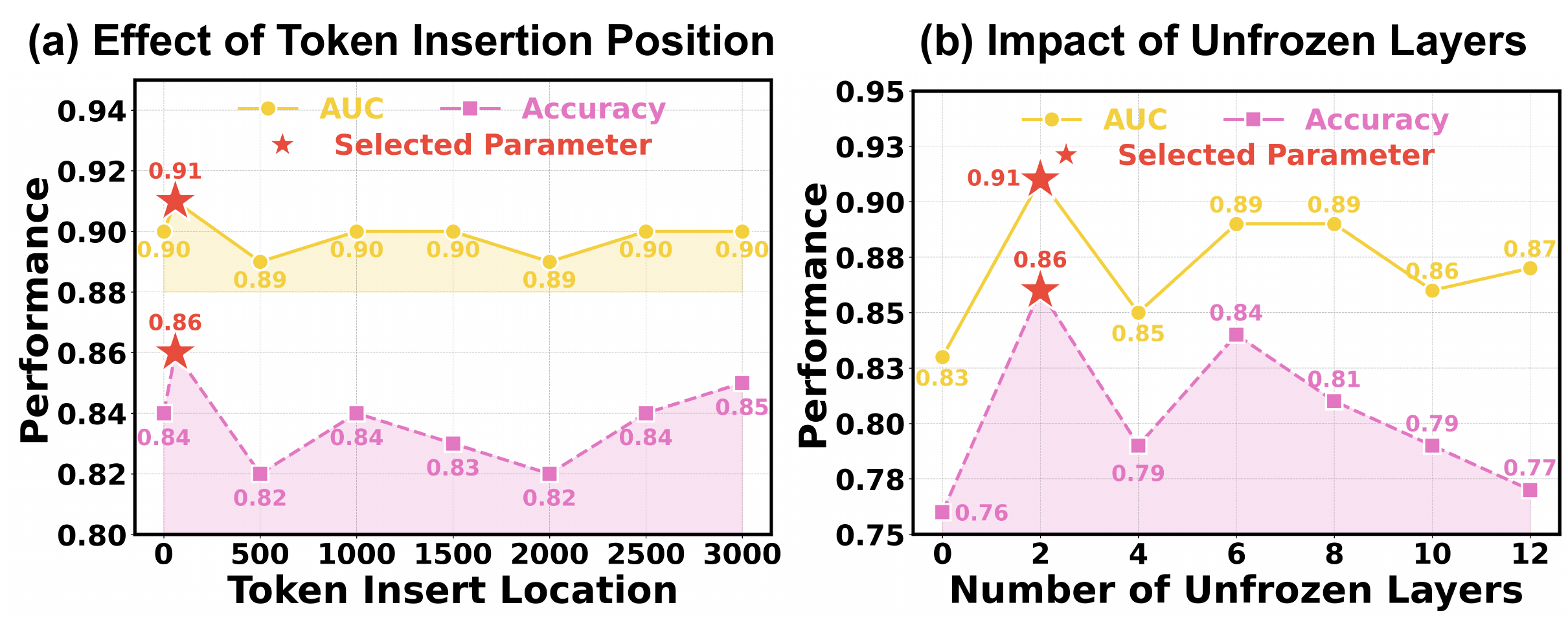}
    \caption{Hyperparameter selection under the trimodal setting in HF diagnosis.}
    \label{fig:hyper}
\end{figure}
\subsection*{5.4. Parameter Analysis}
Furthermore, the analysis was extended to assess the sensitivity of two key parameters: the insertion location of the learnable guidance tokens and the number of unfrozen layers in the foundation models (see Fig.~\ref{fig:hyper}). The trimodal setting for HF diagnosis was used as a typical example for this analysis. The results indicated that model performance remained relatively stable to the insertion location, showing only minor variations when the tokens were placed at the early, middle, or late positions within the human-defined sequence. In contrast, increasing the number of unfrozen layers to two improved downstream task performance up to a point; whilst further increases led to reduced performance followed by fluctuating results.

\begin{figure}[t!]
    \centering
    \includegraphics[width=1\linewidth]{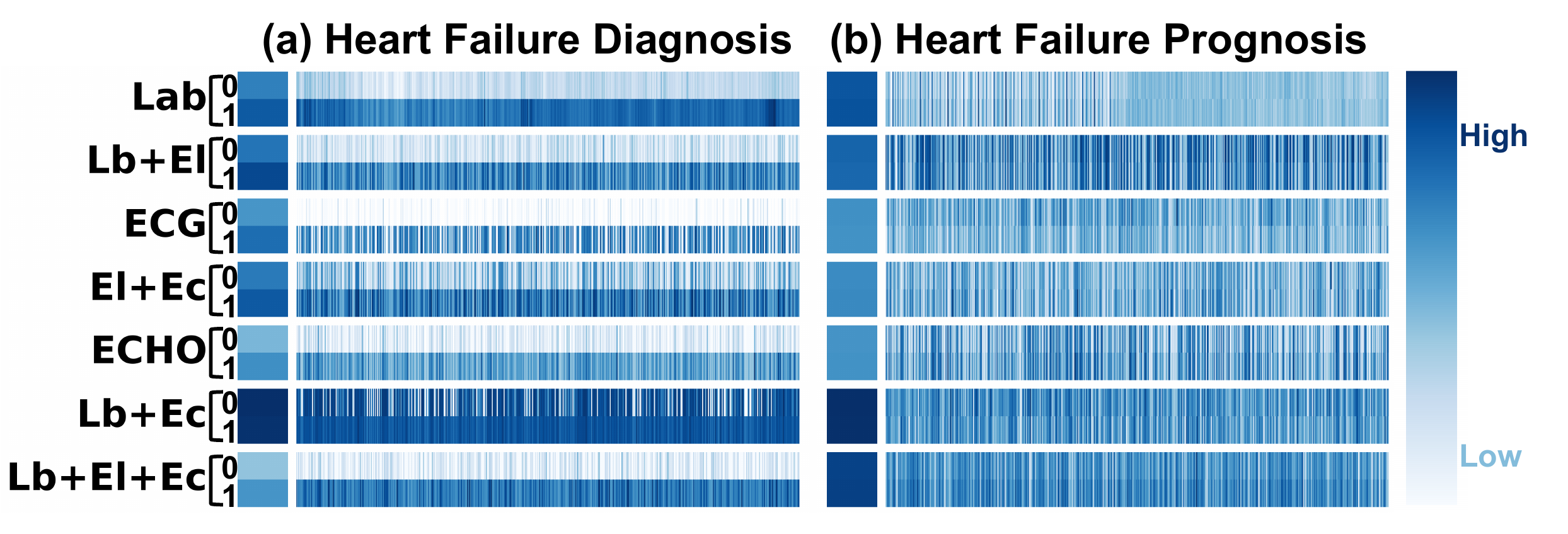}
    \caption{Attention matrix visualization (trimodal): Visualization of modality-specific and shared attention patterns for (a) HF prediction: prevalent (0) vs. non-prevalent (1); and (b) risk prediction: low risk (0) vs. high risk (1). Note that Lb/Lab means laboratory test results; El/ECG means electrocardiograms; Ec/ECHO means echocardiograms.}
    \label{fig:amap}
\end{figure}
\subsection*{5.5. Model Explainability}
To assess the effectiveness of the proposed feature fusion mechanism in downstream predictions, an attention matrix is illustrated to feature contributions across modalities. To further assist medical professionals in understanding the decision-making process of the model, this study systematically identified key features across multiple modalities and elucidated their synergistic contributions. Specifically, a SHAP framework was employed for text-based explanations, while Grad-CAM++ was applied to ECGs and ECHOs  to identify the combined features driving the predictions.  

\medskip
\noindent\textbf{Attention matrix visualization.} 
Figure~\ref{fig:amap} shows the modality-specific and modality-shared attention patterns in the trimodal setting, highlighting the key factors driving to both HF prevalence identification (Fig.~\ref{fig:amap}(a)) and risk stratification (Fig.~\ref{fig:amap}(b)). These figures indicate that, for both tasks, both modality-specific and modality-shared features were attended to, and confirmed that each modality provided distinct but complementary physiological characteristics, which support the clinical reliance on multimodal cardiac assessments. Furthermore, for both tasks, modality-shared features received greater attention, with the fusion of Labs and ECHOs notably attracting significant focus, while modality-specific features were attended to less frequently, with lab-specific features receiving the highest attention among them. Additionally, the prediction of prevalent HF showed a greater reliance on all feature types compared to the prediction of non-prevalent HF, a trend not evident in the high- and low-risk stratification. 

\begin{figure*}[t!]
    \centering
    \includegraphics[width=1\linewidth]{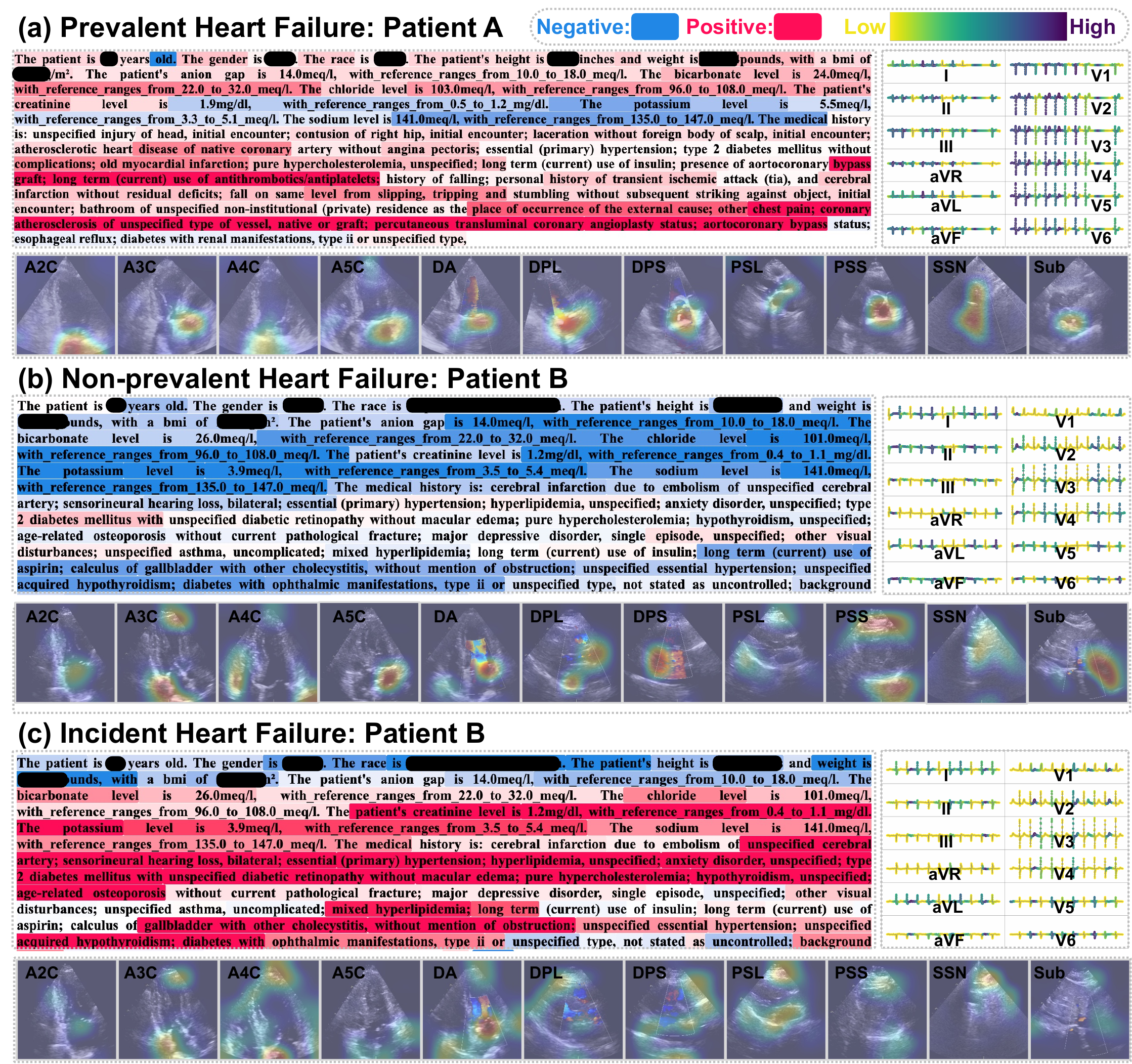}
    \caption{Model Explainability (trimodal): Key features in (a) prevalent HF prediction; (b) non-prevalent HF; (c) follow-up prediction as high risk. In each subplot, SHAP was used to highlight important tokens in text (Labs), where deeper red indicated a stronger positive contribution and deeper blue indicated a stronger negative contribution. Grad-CAM++ was applied to ECGs and ECHOs, where darker regions denoted a stronger influence on the prediction. Note that sensitive information had been marked with black boxes but was used by the model for prediction.}
    \label{fig:explain}
\end{figure*}
\medskip
\noindent\textbf{Prevalent HF prediction.} An example of prevalent HF prediction is illustrated in Fig.~\ref{fig:explain}(a), where tokens such as “antithrombotics,” “antiplatelets,” “chest,” and “pain” appeared as major contributors (darker red indicates stronger impact). In the multimodal context, all 12 ECG leads contributed, with particular emphasis on QRS segments across two consecutive beats. For the echocardiogram, the model attended to clinically significant regions across standard views, including the left atrium and mitral valve in A2C, A4C, and A5C views; the aortic valve in the A3C view; the aortic root in the PLAX view; and the pulmonary artery in the PSAX view. Doppler-based dynamic flow patterns were also captured. The original SHAP output in HTML format and corresponding echocardiographic videos were provided in the Appendix materials. 

\medskip
\noindent\textbf{Non-prevalent HF and incident HF prediction.} To illustrate the capacity of the TGMM to extract task-specific features tailored to different prediction tasks, a patient initially diagnosed with non-prevalent HF but who progressed to HF after 21 months was analyzed. As shown in Fig.~\ref{fig:explain}(b–c), the model relied on distinct features to predict non-prevalent HF and high-risk status, respectively, despite identical input data. Specifically, tokens like “diabetes” and “mellitus” contributed to both tasks, with the latter receiving greater attention. In contrast, terms such as “hypertension” and “hyperlipidemia” were less relevant for diagnosis but significant for high risk prediction. For ECGs, diagnosis primarily relied on the features of two consecutive QRS-waves, while follow-up risk prediction focused on T-wave features. For echocardiograms, diagnosis often focused on cardiac chambers and valves, whereas follow-up risk prediction in certain cases emphasized myocardial contours (e.g., in the A5C). 
\section{Discussion}
This study proposed TGMM, a unified Textual Guidance Multimodal Fusion framework for multiple cardiac tasks. The findings showed that TGMM effectively integrated diverse modalities without compromising performance. Notably, the combination of laboratory tests, ECGs, and echocardiograms outperformed single- or two-modality models in heart failure–related tasks (Table~\ref{tab:task12}), consistent with clinical practice where comprehensive assessment aims to improve diagnoses. On the myocardial infarction dataset, the combination of Labs and ECGs performed comparably to ECGs alone (Table~\ref{tab:MI}), underscoring the pivotal diagnostic role of ECGs. Importantly, TGMM maintained stable performance as modalities were added, whereas Qwen2.5-VL exhibited a decline (AUC 0.97 vs. 0.96, p $<$ 0.05).

Another key strength of the proposed TGMM was its ability to handle the varying availability of clinical data, a common challenge in practice where not all patients underwent the same set of examinations. This contrasted with models like EchoPrime~\cite{vukadinovic2024echoprime}, which were limited to only two modalities. Comparison experiments further highlighted the advantages of the proposed framework, despite yielding comparable results to larger model (e.g., Janus-Pro) in certain cases. Validation using two independent datasets further confirmed the robustness and effectiveness of the proposed framework.

The findings also highlighted the importance of both modality-specific and modality-shared features in clinical tasks. Notably, the attention matrix visualizations in Fig.~\ref{fig:amap} revealed that both features contribute to the predictions, although modality-specific features received less attention in certain tasks (e.g., ECGs in risk prediction). Despite this, they remained valuable. This observation was validated by comparison with alignment-based fusion techniques, like EchoPrime~\cite{vukadinovic2024echoprime}, which prioritize shared information (smallest difference: ECG–ECHO in HF diagnosis, AUC 0.73 vs. 0.78, p $<$ 0.05). Compared to traditional MeanCat~\cite{steyaert2023multimodal}, the TGMM outperformed various combinations of modalities across tasks. While the fusion of Labs and ECHOs yielded comparable diagnostic performance (p = 0.1), the proposed method showed superior results in risk stratification, underscoring its balanced effectiveness.

Finally, the effectiveness of textual guidance mechanism was evaluated. By combining human-defined textual content with model-learned representations, the proposed approach harmonized human guidance and adaptive optimization, enhancing both transparency and outcomes. In addition, the findings showed that the response module outperformed traditional techniques, such as the linear prober, for downstream prediction. This improvement could be attributed to its inherent mechanism of comparing candidate outputs for each option, as well as its superior ability to capture complex contextual relationships within fused features, which was less readily achieved by the linear prober (Table.~\ref{tab:ab}). 

\medskip
\noindent\textbf{Limitation.} While the TGMM showed competitive performance across multiple tasks, several limitations remained. First, its reliance on relatively small public datasets may restrict generalizability, although foundation models had been employed to enhance representation learning and compensate for data scarcity. Second, missing values, a common challenge in medical datasets, introduced additional complexities. While preprocessing techniques, such as transforming tabular data into a language format, might alleviate the impact of missing data, comprehensive datasets still remain essential for optimal performance. Finally, the practical applicability of the framework was constrained by the lack of multi-centre validation.  In light of these challenges, future research could include the generation of high-quality synthetic data to further enhance model generalizability across diverse clinical environments.

\section{Conclusion}
This study introduced a TGMM framework supporting unimodal, bimodal, and trimodal inputs for diverse clinical tasks. Experiments on the curated HFTri-MIMIC dataset demonstrated that TGMM reflected real-world clinical practice by leveraging multiple modalities to improve diagnostic and prognostic accuracy, with external validation further confirming its robustness. Explainability analyses revealed the critical roles of both modality-specific and shared features in clinical data, as proposed in this work, while explainable AI techniques offered transparent feature attributions that shed light on cross-modal interactions. In summary, this study systematically explored multimodal data curation, framework design, and explainability, underscoring the potential of the proposed framework to advance clinical decision support systems.

\bibliographystyle{elsarticle-harv} 
\bibliography{main} 
\end{document}